\documentclass[11pt]{article}

\usepackage[final]{acl}
\usepackage{amsmath, amssymb, amsthm}
\usepackage{subcaption} 
\usepackage{times}
\usepackage[table]{xcolor}
\usepackage{latexsym}
\usepackage{booktabs} 
\usepackage{multirow} 
\usepackage{amsmath} 
\usepackage{nccmath} 
\usepackage{mathtools}
\usepackage{tcolorbox}
\usepackage{tcolorbox}
\usepackage{xcolor}
\tcbuselibrary{skins, breakable}
\usepackage{multirow, booktabs} 
\usepackage[T1]{fontenc}
\usepackage{enumitem}
\usepackage[utf8]{inputenc}

\usepackage{microtype}

\usepackage{inconsolata}

\usepackage{graphicx}

\title{DICA: Dual-Indicator Guided Contrastive Alignment in Multimodal Large Language Models}

\author{
Hao Yang, Jin Wang\thanks{Corresponding author.}\and Xuejie Zhang \\
School of Information Science and Engineering \\
Yunnan University\\ Kunming, China \\
\texttt{yanghao888@stu.ynu.edu.cn},
\texttt{\{wangjin, xjzhang\}@ynu.edu.cn}
}

\begin{document}
\maketitle
\begin{abstract}
Human visual reasoning typically follows a coarse-to-fine attention process, starting from global scene understanding and gradually focusing on question-relevant regions. However, multimodal large language models may deviate from this pattern due to attention drift and the underutilization of visual evidence, which can lead to hallucinations. To mitigate these issues, this study proposes a Dual-Indicator Guided Contrastive Alignment (DICA), which tracks two information-theoretic indicators during inference: Visual Attention Entropy (VAE), which reflects the concentration of visual attention, and Output Image Correlation (OIC), which measures the dependence of generated outputs on the visual input. An abnormal increase in VAE or a decrease in OIC corresponds to different failure modes, which trigger targeted contrastive alignment to restore visual grounding. Experimental results across multiple benchmarks demonstrate that DICA consistently outperforms existing approaches and substantially reduces hallucinations, highlighting the effectiveness of indicator-driven intervention in improving multimodal inference reliability. The code is publicly available at \url{https://github.com/BGWH123/DICA/}.

\end{abstract}

\section{Introduction}

For Visual Question Answering (VQA), human visual attention follows a coarse-to-fine paradigm.  typically begin by conducting a global scan of the scene to acquire an overall semantic structure and contextual cues. As the question is progressively interpreted, attention gradually narrows and shifts toward the local regions most relevant to the query, and the final answer becomes highly dependent on these key visual details \cite{NAVON1977353}. This transition from global context awareness to fine-grained semantic focus has been repeatedly validated in cognitive and vision studies \cite{das2017human,yarbus2013eye,itti2001computational}.

To investigate whether modern Multimodal Large Language Models (MLLMs) exhibit a similar attention pattern, a systematic empirical analysis is conducted in Section~\ref{motivation}. The results reveal a consistent transition from global attention to localized focus. As reasoning unfolds, the MLLMs' visual attention progressively shifts from a diffuse global distribution to a concentrated focus on regions semantically aligned with the query.

\begin{figure}[t]
    \centering

    \includegraphics[width=0.9\linewidth]{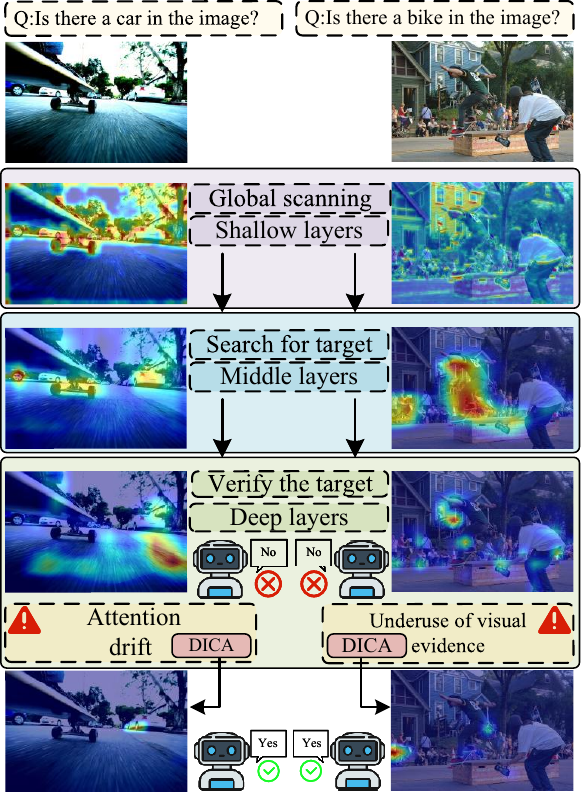}
    \caption{An example of attention drift and underuse of visual evidence.}
    \label{fig:two_question}
\end{figure}

Besides, two major issues are found in the experiments, as shown in Figure \ref{fig:two_question}. \textbf{(a) Attention Drift:} During reasoning, the model's visual attention may suddenly disperse or shift to regions that are semantically irrelevant to the query. \textbf{(b) Underuse of Visual Evidence}: The model's reliance on visual cues does not consistently escalate as the network deepens. Conversely, in some cases, the reliance on visual content even degrades in deeper layers. These two issues consequently cause MLLMs to yield incorrect response.

However, existing works, including both training-based approaches~\cite{cai2024vip,villa2025eagle,liu2023mitigating,yu2024rlhf} and training-free inference-time methods~\cite{leng2024mitigating,huang2024opera,favero2024multi,woo2025don,chen2025mixture}, generally treat hallucination as a monolithic phenomenon and lack explicit mechanisms to diagnose its root causes.
They fail to distinguish whether hallucinations arise from attention drift or from underuse of visual evidence, and therefore apply uniform correction strategies that may be misaligned with the actual failure modes. As a result, these methods are often unable to perform precise, adaptive interventions when different hallucination patterns occur during reasoning.


In contrast, we propose Dual-Indicator Guided Contrastive Alignment (DICA), which explicitly diagnoses and addresses different hallucination sources during inference. Specifically, DICA continuously monitors two typical failure patterns, attention drift and underuse of visual evidence, via dedicated indicators. Once either indicator exceeds a predefined threshold, a targeted correction is triggered. If attention drift is detected, the hidden states of a neutral query are incorporated to calibrate the current hidden states; if underuse of visual evidence is detected, the original prediction is calibrated using the best alternative distribution.

Extensive experiments on multiple benchmarks show that DICA consistently enhances reasoning accuracy, validating both our theoretical analysis and the practical efficacy of the proposed approach.




The contributions are summarized as follows:
\begin{itemize}
    \item We empirically uncover two key bottlenecks about MLLM hallucinations: \textbf{(a) attention drift} and \textbf{(b) underuse of visual evidence}.
    \item DICA is proposed as a method that can be applied to different models and performs distribution correction at inference time, without requiring retraining or changes to the model architecture.
    \item Extensive evaluations demonstrate that DICA significantly boosts reasoning accuracy and suppresses hallucinations by effectively strengthening the model's reliance on visual evidence.
\end{itemize}

\section{Motivation}
\label{motivation}


\subsection{Indicator Definition}
\label{definition}
To analyze whether MLLMs exhibit attention patterns similar to human visual attention and whether their answers remain grounded in visual cues, we introduce two diagnostic metrics\footnote{The theoretical basis and mathematical proofs for these metrics are provided in Appendices~\ref{sec:appendixa} and~\ref{sec:appendixb}, respectively.}: \textbf{(1) Visual Attention Entropy (VAE)}, which captures the concentration level of visual attention, and \textbf{(2) Output--Image Correlation (OIC)}, which measures the model's reliance on visual cues. These metrics serve as diagnostic indicators for identifying abnormal attention and evidence-utilization behaviors rather than causal explanations of hallucination.

\noindent\textbf{Visual Attention Entropy.}  
Shannon entropy \cite{6773024} is employed to quantify the concentration of attention over visual tokens.
Let the attention weights over all tokens be $\{ a^{(l,t)}_i \}_{i=1}^{T}$ at layer $l$ and decoding step $t$, 
and let $m \in \{0, 1\}^T$ denote the image token mask, where $m_i = 1$ indicates a visual token. 
The attention weights corresponding to visual tokens are first selected:
\begin{equation}
\mathcal{A}^{(l,t)} = \{ a^{(l,t)}_i \mid m_i = 1 \}.
\end{equation}
These are then normalized to form a probability distribution:
\begin{equation}
p^{(l,t)}_i = \frac{a^{(l,t)}_i}{\sum_{j: m_j = 1} a^{(l,t)}_j}, 
\quad i \in \{ j \mid m_j = 1 \}.
\end{equation}
The VAE for layer $l$ at step $t$ is computed as
\begin{equation}
\mathrm{VAE}^{(l,t)} = - \sum_{i: m_i = 1} p^{(l,t)}_i \log p^{(l,t)}_i.
\end{equation}
The sequence $ \{\text{VAE}^{(l,1)},\dots, \text{VAE}^{(l,T)}\}$ represents the evolution of the model's visual attention across decoding steps for each layer and token.  
Higher entropy indicates more dispersed attention, while lower entropy indicates more concentrated attention over a subset of visual tokens.

\noindent\textbf{Output--Image Correlation.} OIC measures the influence of image-conditioned hidden states on the model's output distributions relative to unconditioned hidden states.

Let $h^{(l,t)}$ denote the hidden states at decoding step $t$ and layer $l$ when conditioned on both text and image, 
and let $h^{(l,t)}_{\text{wi}}$ denote the hidden states when conditioned only on text, without image.
The corresponding output logits are obtained via a linear projection:
\begin{align}
\mathbf{z}^{(l,t)} = \mathbf{h}^{(l,t)} W^\top,  \\
\mathbf{z}^{(l,t)}_{\text{wi}} = \mathbf{h}^{(l,t)}_{\text{wi}} W^\top,
\end{align}
\noindent where $W$ is the language modeling head parameters matrix.  
The probability distributions over the vocabulary are then computed using the softmax function:
\begin{align}
\mathbf{p}^{(l,t)} = \text{softmax}(\mathbf{z}^{(l,t)}), \\
\mathbf{p}^{(l,t)}_{\text{wi}} = \text{softmax}(\mathbf{z}^{(l,t)}_{\text{wi}}).
\end{align}
The OIC for each token is defined:
\begin{equation}
\text{OIC}^{(l,t)} = \sum_{v} p^{(l,t)}_v \left( \log p^{(l,t)}_v - \log p^{(l,t)}_{\text{wi},v} \right),
\end{equation}
\noindent where $v$ indexes the vocabulary.  
The sequence $\{ \text{OIC}^{(l,1)}, \text{OIC}^{(l,2)}, \dots, \text{OIC}^{(l,T)} \}$ then represents the evolution of the model's output dependence on image information across decoding steps for each layer and token.  
Higher OIC values indicate stronger reliance on visual input, while lower values indicate that the output primarily depends on text context.

\subsection{Findings}
\newtcolorbox{insightbox}[1]{
    enhanced,
    sharp corners,
    colback=gray!10,
    colframe=gray!50!black,
    boxrule=0.4pt,               
    top=5pt, bottom=5pt,         
    left=6pt, right=6pt,         
    boxsep=0pt,                  
    attach boxed title to top left={
        yshift=-1.2mm,           
        xshift=2mm               
    },
    boxed title style={
        colback=gray!70!black,
        sharp corners,
        size=small,
        boxrule=0pt,
        top=1.5pt, bottom=1.5pt,
        left=4pt, right=4pt      
    },
    coltitle=white,
    fonttitle=\normalsize\normalfont, 
    title={#1},
    before skip=4pt,             
    after skip=4pt               
}

\begin{figure}[t]
    \centering
    \includegraphics[width=0.9\linewidth]{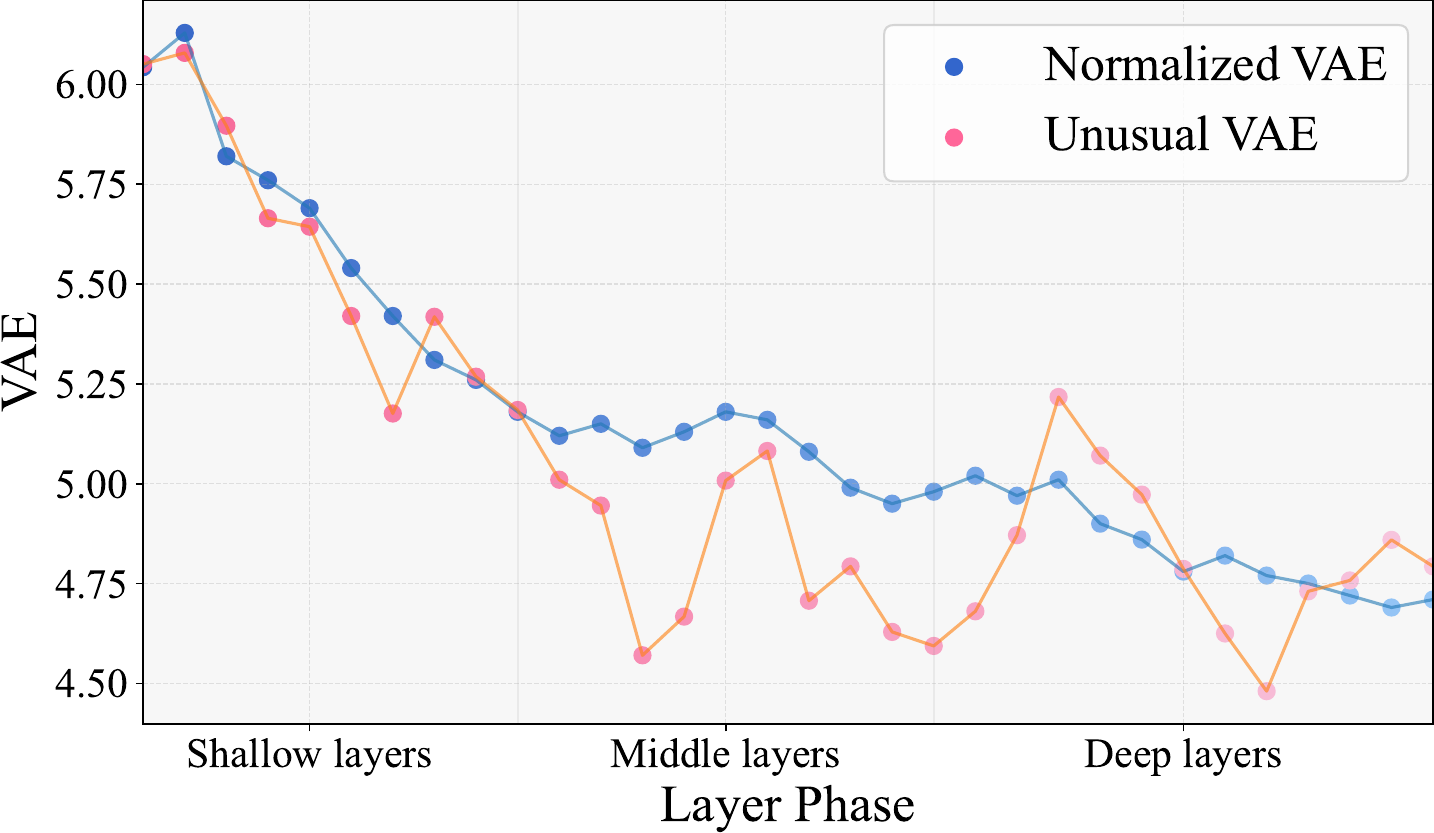}\\[2mm]
    \includegraphics[width=0.9\linewidth]{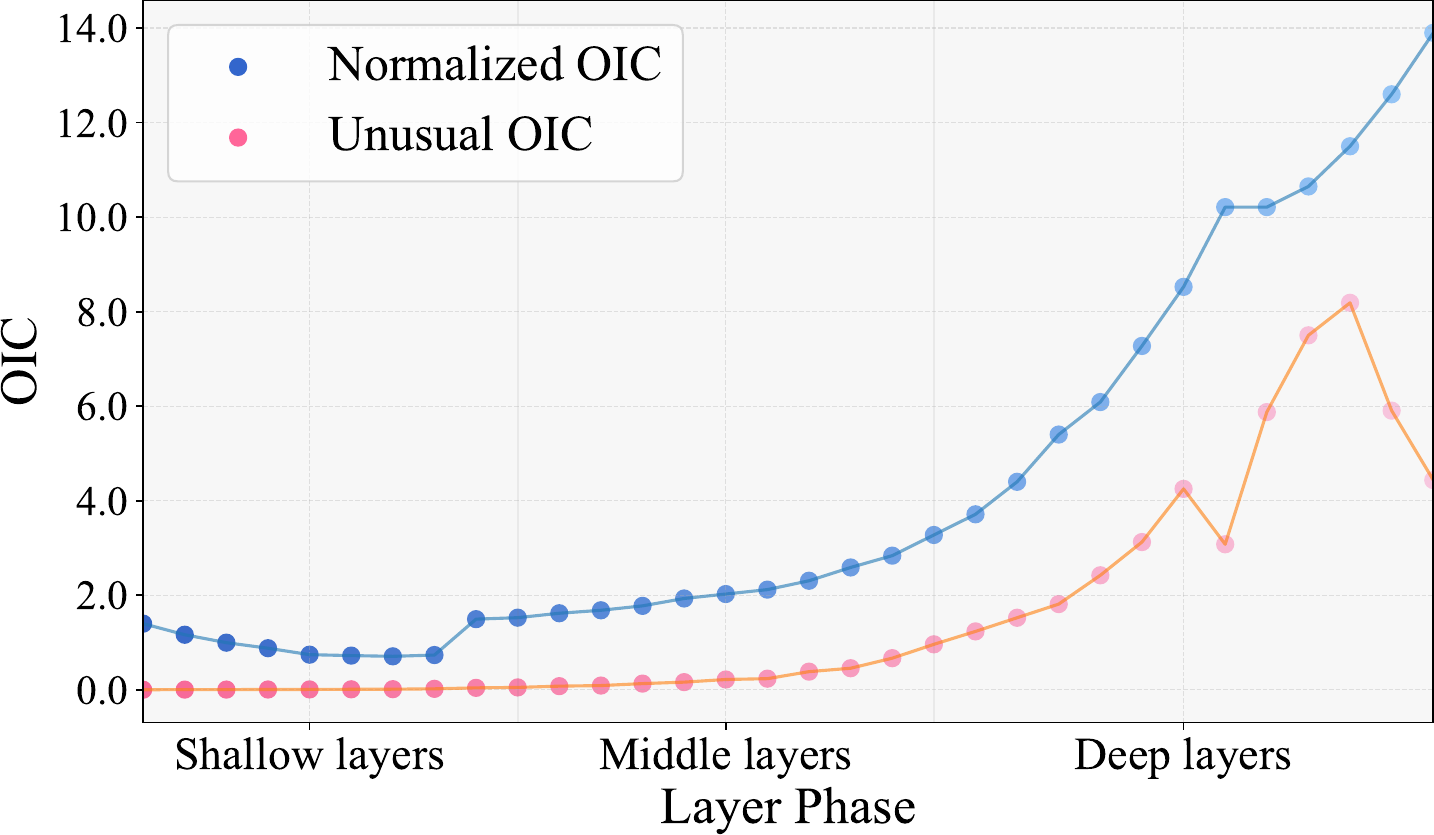}
\caption{Top: Evolution of VAE across layer phases. Bottom: Evolution of OIC across layer phases.}
    \label{fig:vae_and_oic_1}
\end{figure}

\begin{figure*}[t]
    \centering

    \includegraphics[width=1.0\linewidth]{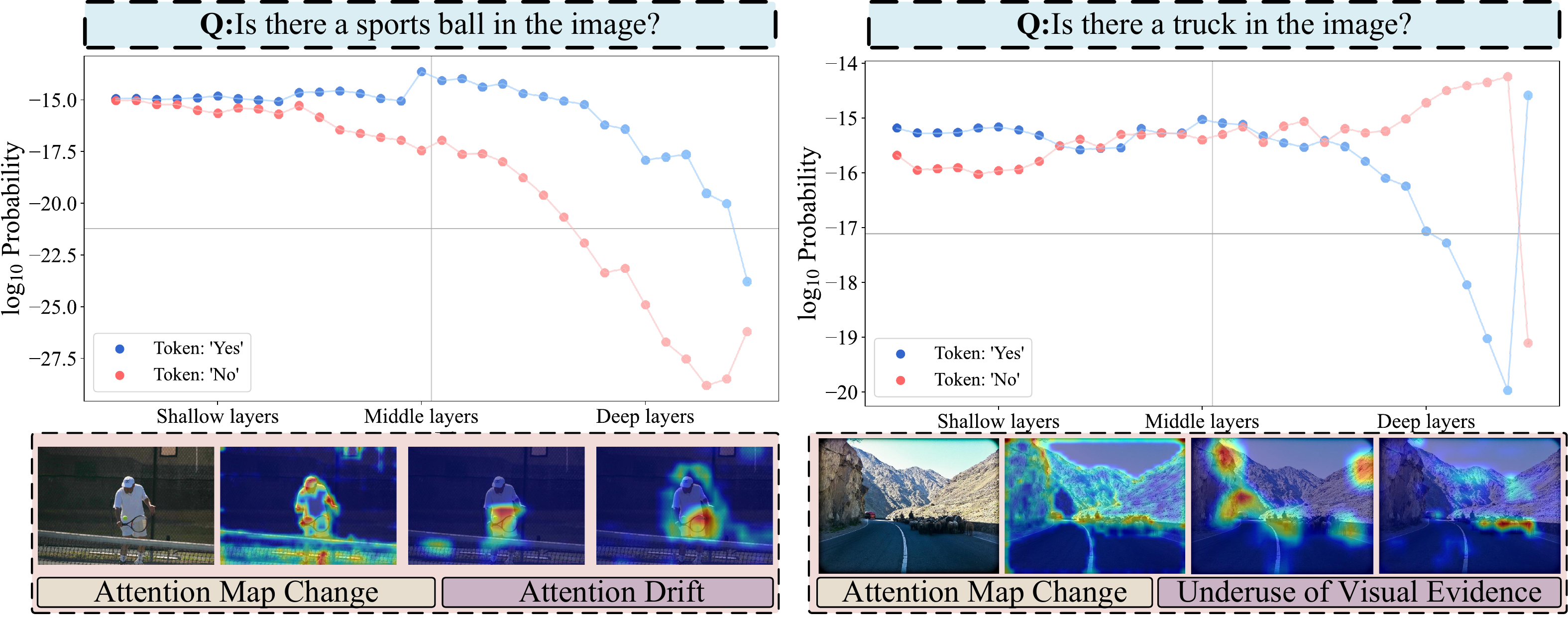}
    \caption{Evolution of answer log-probabilities and attention maps across layers.}
    \label{fig:Adver_results}
\end{figure*}

\begin{insightbox}{Insight 1}
Visual attention exhibits a transition from diffuse to more concentrated patterns, as reflected by a generally decreasing VAE. 
\end{insightbox}
We systematically study cross-modal attention in LLaVA-1.5 \cite{liu2024improved} across transformer layers. 
Specifically, we isolate the attention tokens aligned with visual tokens and calculate their VAE to measure the degree of focus or dispersion in the attention distribution.
As shown at the top of Figure \ref{fig:vae_and_oic_1}, it is observed that the downward trajectory of VAE largely supports hypothesis of a global-to-local attention evolution.  

However, during decoding, VAE occasionally shows sharp increases. These peaks indicate sudden rises in visual attention entropy, meaning the model's attention becomes unexpectedly dispersed. Such behavior typically signals confusion or a disconnect from visual evidence and may foreshadow upcoming hallucinated content, suggesting that focused visual cues no longer support the generated output. This anomaly will be further discussed in \ref{averse results}.
\begin{insightbox}{Insight 2}
During the reasoning process, the model progressively increases its reliance on visual information, as evidenced by rising OIC values. 
\end{insightbox}
The model's reliance on visual information was further quantified by the OIC metric, as illustrated in the bottom part of Figure~\ref{fig:vae_and_oic_1}. Under normal reasoning, OIC values typically show a gradual increase or remain high, indicating that the model can consistently and effectively leverage cross-modal mappings to incorporate image features when constructing its answers, thereby ensuring that the generated text is grounded in visual evidence. 

However, at certain moments, the metric exhibits abnormal sudden drops. Such abrupt decreases signal a sharp reduction in the model's dependence on visual input, with the generation distribution becoming primarily driven by the internal priors of the language model. This sudden disengagement from the visual context often implies that the model is performing blind inference based solely on its textual priors, which can easily lead to hallucinated outputs that are inconsistent with the image content. This anomaly will be discussed in Section \ref{averse results}.

\subsection{Anomaly Analysis}
\label{averse results}

In the preceding analysis of indicators, two types of abnormal phenomena have been identified: \textbf{sudden increases in VAE and drops in OIC}. To further investigate the consequences of these two situations, a step-by-step visualization of the reasoning trajectories in which such abnormalities occur is conducted. This includes examining layer-wise changes in attention heatmaps as well as the evolution of the answer log-probabilities, as illustrated in Figure~\ref{fig:Adver_results}. After correction by DICA, the results are provided in \ref{Effectiveness of DICA in Visible}.

When a sudden surge in VAE is observed, the visual attention previously concentrated on relevant regions abruptly disperses or drifts toward areas semantically irrelevant to the question. Such attention drift interrupts the expected global-to-local focusing process and prevents the model from maintaining stable reliance on critical visual evidence in subsequent layers. This phenomenon is referred to as \textbf{attention drift}. The corresponding answer probability curves also exhibit pronounced fluctuations, indicating that the model progressively deviates from a visually grounded reasoning trajectory and becomes prone to erroneous or hallucinatory predictions. The left of Figure~\ref{fig:Adver_results} shows a representative example of this phenomenon.

In contrast, when OIC experiences a decline, the output distribution reflects a clear reduction in reliance on image information. In this situation, the generation process becomes increasingly dominated by language priors, causing the model to ignore visual evidence and shift toward prior-driven inference, i.e., \textbf{underuse of visual evidence}. During this phase, the answer become unstable and decline, eventually diverging from the actual visual content. As shown in the right of Figure~\ref{fig:Adver_results}, this lack of visual grounding similarly leads to incorrect predictions.


\begin{figure*}[t]
    \centering

    \includegraphics[width=1.0\linewidth]{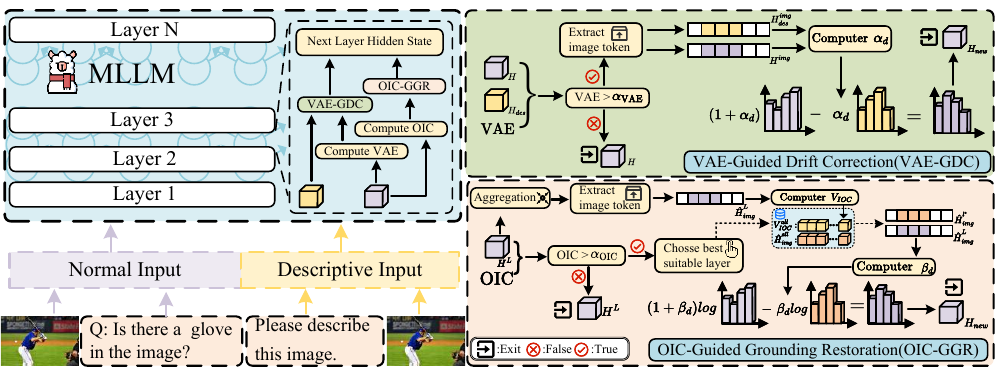}
    \caption{The overall architecture of DICA.}
    \label{fig:main}
\end{figure*}
\section{Dual-Indicator Guided Alignment}

Based on the above analysis, the rise in VAE and the drop in OIC lead to attention drift and loss of visual grounding, which in turn cause reasoning errors. Thus, a suitable method should address these issues and maintain the model's alignment with visual evidence. Therefore, the DICA architecture is proposed, as shown in Figure~\ref{fig:main}. It is composed of two components, namely VAE-Guided Drift Correction and OIC-Guided Grounding Restoration.


At each MLLM layer $l$ and decoding step $t$, the two indicators are computed:
\begin{align}
\mathrm{VAE}^{(l,t)}&=\mathrm{VAE}(a^{(l,t)}), \\
\mathrm{OIC}^{(l,t)}&=\mathrm{OIC}(\mathbf{h}^{(l,t)},\mathbf{h}^{(l,t)}_{\text{wi}}),
\end{align}
\noindent where both functions follow the definitions provided in Section~\ref{definition}. 

\subsection{VAE-Guided Drift Correction}
Attention drift is identified when the VAE surpasses a predefined threshold corresponding to a proportion $\alpha_{\mathrm{VAE}}$ of its dynamic range, indicating dispersion of visual attention at the feature level. In this state, the hidden representations of image tokens in the Normal path are contaminated by high-entropy noise, causing the model to disperse focus toward irrelevant regions. 

To correct this feature-level deviation, we leverage the Descriptive path. Due to its inherently descriptive generation process, the Descriptive path captures richer fine-grained details \cite{li2023blip,peng2023kosmos,liu2023visual} and maintains a more uniform attention distribution across regions. Based on this property, we perform contrastive decoding directly on hidden states to mitigate the effect of high-entropy contamination.

Specifically, let \(H^{\mathrm{img}}\) and \(H^{\mathrm{img}}_{\mathrm{desc}}\) denote the image-token hidden states in the Normal and Descriptive paths, respectively. We map them to logits \(Z\) and \(Z_{\mathrm{desc}}\) via the language modeling projection layer. Then, a dynamic weight is computed \(\alpha_d\) to control the correction strength based on their distributional discrepancy: \begin{equation} \alpha_d = \operatorname{clip}\Big( 1 - \mathrm{JSD}(Z \,\|Z_{\mathrm{desc}}), 0, 1 \Big), \end{equation}

\noindent where \(\mathrm{JSD}(\cdot)\) measures the Jensen--Shannon Divergence~\cite{61115} between the hidden-state feature distributions. The corrected hidden state is then computed as:
\begin{equation}
H^{\mathrm{new}} = (1+\alpha_d)\, H^{\mathrm{img}} - \alpha_d \, H^{\mathrm{img}}_{\mathrm{desc}}.
\end{equation}

\subsection{OIC-Guided Grounding Restoration}

For step \(t\) of a given sample \(b\), if the OIC value shows a persistent decrease and falls below a threshold defined as a proportion \(\alpha_{\mathrm{OIC}}\) of its own range, we consider this an indication of underuse of visual evidence.

After computing OIC, we further require a metric that explicitly captures whether the model attends to image regions during reasoning. To this end, we define the image token attention vector \(V_{\mathrm{OIC}}\), which measures the explicit attention strength that the current language token allocates to the image tokens.

Formally, let \(\mathbf{A}\in\mathbb{R}^{B\times H\times T\times T}\) denote the multi-head attention scores of a given layer \(l\), where \(B\) is the batch size, \(H\) is the number of attention heads, and \(T\) is the length of the input sequence. Let \(\mathbf{m}\in\mathbb{R}^{B\times T}\) be a binary mask indicating the positions of the image tokens within the sequence. At the decoding step \(t\), the model uses the last query position to attend to all key positions. Following prior findings \cite{neo2402interpreting,xu2025mitigating} that the maximum activation across heads provides a reliable estimate of the model's dominant focus, \(\mathbf{A}\) is aggregated across the head dimension as:
\begin{equation}
\hat{\mathbf{A}}_{b,t,j}
=
\max_{h\in\{1,\dots,H\}}
\mathbf{A}_{b,h,t,j},
\hat{\mathbf{A}}\in\mathbb{R}^{B\times T\times T},
\end{equation}

\noindent where \(j\) indexes the source positions attended to.

The aggregated attention distribution is then restricted to the positions corresponding to image tokens, as specified by the mask \(\mathbf{m}\). For each sample \(b\), decoding step \(t\) and  layer \(l\), the raw image-directed attention vector is
\begin{equation}
\mathbf{v}_{b,t}
=
\left[
\hat{\mathbf{A}}_{b,t,j}
\;\middle|\;
\mathbf{m}_{b,j}=1
\right].
\end{equation}
To obtain a normalized attention profile over image tokens, we apply a softmax operation:
\begin{equation}
V_{\mathrm{OIC}}^{(b,t)}
=
\operatorname{softmax}
\left(
\mathbf{v}_{b,t}
\right),
\end{equation}
where \( V_{\mathrm{OIC}}^{(b,t)} \) represents the relative attention probability that the model allocates to the image token at the current decoding step in the \( l \)-th layer. For simplicity, we denote it as \( V_{\mathrm{OIC}}^{l} \).


The JSD  is then computed between this layer and each preceding layer to quantify the layer-wise difference in visual attention distributions.
\begin{equation}
\mathrm{dist}(V_{\mathrm{OIC}}^{N}, V_{\mathrm{OIC}}^{L}) = \mathrm{JSD}(V_{\mathrm{OIC}}^{N} \,\|\,V_{\mathrm{OIC}}^{L}),
\end{equation}

\noindent where \(L\) denotes the final decoding layer. The layer $N = 1,\dots, L-1$. Exhibiting the maximal divergence is selected as the target intermediate layer $l^*$ :
\begin{equation}
l^* = \mathop{\arg\max}\limits_{\mathclap{N=1,\dots,L-1}}
\mathrm{dist}\big(V_{\mathrm{OIC}}^{(N)},V_{\mathrm{OIC}}^{(L)}\big).
\end{equation}
To adaptively modulate the strength of grounding restoration, we define a dynamic scaling factor \(\beta_t\) based on the inter-layer divergence of visual attention distributions:
\begin{equation}
\beta_d
=
\operatorname{clip}
\Big(
\mathrm{JSD}
\big(
V_{\mathrm{OIC}}^{l^*}
\;\|\;
V_{\mathrm{OIC}}^{L}
\big),
\;
0,
1
\Big),
\end{equation}
\noindent 
Once the layer \(l^*\) is selected, intermediate-layer contrastive decoding is performed by comparing the predicted token distributions from the final layer and the selected intermediate layer. Let \(H^{L}_{\mathrm{img}}\) and \(H^{l^*}_{\mathrm{img}}\) denote the vocabulary logits produced by the final and intermediate layers under image-aware hidden representations. The contrastive decoding distribution is then computed as:
\begin{equation}
H_{\mathrm{new}}
=
\medmath{
\mathrm{softmax}\Big(
(1+\beta_d)\log H^{L}_{\mathrm{img}}
-
\beta_d \log H^{l^*}_{\mathrm{img}}
\Big)
}.
\end{equation}


\definecolor{lightblue}{RGB}{230,240,255}
\definecolor{lightpink}{RGB}{255,230,240}
\definecolor{lightgreen}{RGB}{230,255,240}
\begin{table*}[t]
\setlength{\tabcolsep}{3pt}
\renewcommand{\arraystretch}{0.95}
\small
\centering

\begin{tabular}{llrrrrrrrrrrrr}
\toprule
Setting & Method &
\multicolumn{4}{c}{MSCOCO} &
\multicolumn{4}{c}{AOKVQA} &
\multicolumn{4}{c}{GQA} \\
\cmidrule(lr){3-6}\cmidrule(lr){7-10}\cmidrule(lr){11-14}

& & 
\multicolumn{2}{c}{LLaVA-v1.5} & \multicolumn{2}{c}{Qwen-VL} &
\multicolumn{2}{c}{LLaVA-v1.5} & \multicolumn{2}{c}{Qwen-VL} &
\multicolumn{2}{c}{LLaVA-v1.5} & \multicolumn{2}{c}{Qwen-VL} \\
\cmidrule(lr){3-4}\cmidrule(lr){5-6}\cmidrule(lr){7-8}\cmidrule(lr){9-10}\cmidrule(lr){11-12}\cmidrule(lr){13-14}

& & Acc & F1 & Acc & F1 & Acc & F1 & Acc & F1 & Acc & F1 & Acc & F1 \\
\midrule

\multirow{8}{*}{Random}
& sample  & 83.80 & 84.20 & 84.90 & 82.90 & 81.80 & 83.50 & 86.80 & 85.80 & 81.60 & 83.50 & 83.30 & 79.20 \\
& AvisC   & 82.30 & 83.50 & 82.90 & 80.00 & 79.10 & 82.10 & 84.70 & 83.00 & 79.00 & 82.20 & 80.50 & 77.80 \\
    & VCD      & 85.00 & 84.20 & 85.50 & 83.60 & 81.20 & 83.20 & 87.40 & 86.60 & 82.20 & 84.10 & 82.00 & 80.50 \\
& OPERA    & 89.20 & 88.81 & \underline{87.31} & \textbf{86.92} & \underline{88.02} & 84.59 & \underline{88.19} & \underline{88.43} & \underline{88.13} & \textbf{88.91} & 86.02 & 85.29 \\
& M3ID    & 86.10 & 85.00 & 85.30 & 83.40 & 82.90 & 84.60 & 87.10 & 85.90 & 83.30 & 84.50 & 82.40 & 79.70 \\
&MoD &\underline{89.20} &\textbf{89.10}&86.00&84.10&86.50&\underline{87.50}&87.80&86.90&86.20&87.40&83.80&82.30\\
\rowcolor{lightblue}
& DICA (ours)      &\textbf{89.23} & \underline{88.82} & \textbf{87.37} & \underline{85.97} & \textbf{89.23} & \textbf{89.42} & \textbf{89.47} & \textbf{88.93} & \textbf{89.09} & \underline{88.65} & \textbf{86.67} & \textbf{85.42} \\
\rowcolor{lightpink}
&\small{\textit{w/o} VAE\text{-}GDC} & 86.50 & 85.04 & 86.01 & 84.19 &88.60 & 87.78& 87.23 & 86.48 & 88.40 & 88.06 & 83.57 & 81.93 \\
\rowcolor{lightgreen}
& \small{\textit{w/o} OIC\text{-}GGR}   & 87.76 & 86.67 & 86.13& 84.24 &88.90 & 89.07 & 89.15 & 88.58 & 88.57 & 88.11 & 86.23 & 84.77 \\

\midrule

\multirow{8}{*}{Popular}
& sample  & 82.00 & 82.60 & 84.00 & 81.90 & 75.30 & 78.70 & 85.60 & 84.70 & 73.10 & 77.50 & 75.90 & 74.90 \\
& AvisC   & 78.20 & 80.50 & 82.80 & 80.10 & 71.80 & 77.20 & 83.90 & 83.40 & 67.40 & 74.80 & 74.20 & 72.30 \\
& VCD      & 82.10 & 83.20 & 84.90 & 83.60 & 74.70 & 78.50 & 86.30 & 85.10 & 71.50 & 76.80 & 75.90 & 75.60 \\
& OPERA    & \underline{86.64} & \underline{86.62} & \underline{87.44} & \textbf{86.68} & \underline{83.22} & \underline{84.67} & \underline{87.91} & \underline{87.13} & \underline{79.27} & \underline{82.11} & \underline{81.97} & \textbf{82.12} \\
& M3ID    & 82.80 & 84.10 & 84.20 & 82.70 & 75.80 & 79.40 & 85.90 & 84.60 & 72.30 & 77.10 & 76.80 & 77.00 \\
&MoD &85.70&86.10&85.60&83.70&79.50&82.20&86.50&85.60&74.00&78.60&79.80&78.90\\
\rowcolor{lightblue}
& DICA (ours)      & \textbf{87.72} & \textbf{86.65} & \textbf{87.50} & \underline{86.13} & \textbf{86.64} & \textbf{85.99} & \textbf{88.19} & \textbf{87.50} & \textbf{81.88} & \textbf{82.52} & \textbf{82.13} & \underline{81.72} \\
\rowcolor{lightpink}
&\small{\textit{w/o} VAE\text{-}GDC} & 86.40 & 84.92 & 85.50 & 83.57 & 85.78 & 85.87 & 86.97 & 86.10 & 79.43 & 81.08 & 78.93 &77.90  \\
\rowcolor{lightgreen}
& \small{\textit{w/o} OIC\text{-}GGR}   & 86.53 & 85.06 & 85.63 & 83.75 &86.46 & 85.83  & 87.03 & 86.19 & 81.26 & 82.02 & 79.13 & 78.07 \\
\midrule

\multirow{8}{*}{Adversarial}
& sample  & 75.80 & 78.10 & 82.10 & 80.20 & 67.40 & 73.70 & 80.40 & 80.50 & 68.00 & 74.50 & 75.50 & 74.40 \\
& AvisC   & 74.20 & 77.70 & 81.20 & 78.50 & 64.40 & 73.00 & 78.10 & 77.30 & 64.10 & 72.90 & 75.50 & 73.40 \\
& VCD      & 76.30 & 78.70 & 84.00 & 82.00 & 68.10 & 74.60 & 80.70 & 80.80 & 67.60 & 74.50 & 76.70 & 76.20 \\
& OPERA    & \underline{81.24} & \underline{81.38} & \underline{84.78} & \underline{83.45} & \underline{73.82} & \underline{77.91} & 80.82 & \textbf{81.54} & \underline{75.00} & \underline{78.71} & \textbf{80.24} & \textbf{80.64} \\
& M3ID    & 77.10 & 78.90 & 83.20 & 80.80 & 68.30 & 74.60 & 80.50 & 80.40 & 67.20 & 74.00 & 77.10 & 76.70 \\
&MoD &79.70&81.30&84.00&82.30&69.10&75.40&\underline{81.00}&80.90&68.70&75.30&78.90&78.10\\
\rowcolor{lightblue}
& DICA (ours)      & \textbf{83.20} & \textbf{82.55} & \textbf{84.80} & \textbf{83.67} & \textbf{77.77} & \textbf{78.63} & \textbf{81.33} & \underline{80.98} & \textbf{78.40} & \textbf{79.86} & \underline{79.23} & \underline{78.21} \\
\rowcolor{lightpink}
&\small{\textit{w/o} VAE\text{-}GDC} & 81.40 & 81.70  &83.80 & 81.98 & 76.47 & 76.41 & 81.07 & 80.68 & 77.79 & 79.37 & 78.57 & 78.53 \\
\rowcolor{lightgreen}
& \small{\textit{w/o} OIC\text{-}GGR}   & 82.37 & 81.94& 84.00 & 82.24 & 77.47 & 76.45 & 81.23 & 81.16 &77.53 & 80.41 & 78.80 & 77.89 \\
\bottomrule
\end{tabular}
\caption{Results on POPE benchmark. \textit{Acc} and \textit{F1} denote Accuracy and F1 score, respectively. \textbf{Bold} indicates the best performance, and \underline{underline} denotes the second best.}
\label{tab:performance-transposed}
\end{table*}

\section{Experiments}

\subsection{Experiments Setup}
\textbf{Datasets and Metrics.} We evaluate DICA on three benchmarks as follows:

\noindent\textbf{(1) POPE}~\cite{li-etal-2023-evaluating}. The benchmark is designed to test whether MLLMs can correctly judge the presence or absence of specific objects in an image, thereby identifying object-level hallucinations. It formulates the task as Yes/No queries based on object annotations. The dataset is balanced (50\% real objects and 50\% non-existent objects) and employs three sampling strategies: random, popular, and adversarial. Built from MSCOCO \cite{lin2014microsoft}, A-OKVQA \cite{schwenk2022okvqa}, and GQA \cite{hudson2019gqa}, POPE contains approximately 27k query--answer pairs for evaluation.

\noindent\textbf{(2) MME}~ \cite{zhang2021mme}. The comprehensive benchmark is designed to assess MLLM's performance across multiple dimensions. It covers ten perception-oriented tasks and four cognition-related tasks. Object-level hallucination is examined through subsets evaluating object presence and counting, while attribute-level hallucinations are analyzed via subsets focusing on object color and position.

\noindent\textbf{(3) CHAIR}~ \cite{rohrbach-etal-2018-object}. In this evaluation framework, object hallucination in image captioning is assessed by prompting MLLMs to produce descriptive captions for a randomly chosen subset of 500 images from the MS-COCO validation set. The metric quantifies hallucination by measuring the fraction of objects referenced in the generated caption that are absent from the corresponding ground-truth annotations. It provides two primary metrics: $\text{CHAIR}_i$ at the object-instance level and $\text{CHAIR}_s$ at the sentence level.

\noindent\textbf{Implementation Details.} In experiments, we selected the models LLaVA-1.5 \cite{liu2024improved} and Qwen-VL \cite{bai2023qwen} as the backbones. The hyperparameters are set as $\alpha_{\mathrm{VAE}} = 0.8$, and $\alpha_{\mathrm{OIC}} = 0.5$.\footnote{Hyperparameter analysis is provided in \ref{Hyperparameter analysis}.} All experiments were conducted on an NVIDIA A100 GPU.

\noindent\textbf{Baselines.} All baseline methods in our experiments are selected from VCD \cite{leng2024mitigating}, OPERA \cite{huang2024opera}, M3iD \cite{favero2024multi}, AvisC \cite{woo2025don}, and MoD \cite{chen2025mixture}. These methods aim to mitigate hallucinations by reducing reliance on language priors during the decoding process.
 

\subsection{Main Results}

\noindent\textbf{Results on POPE.}  
Table~\ref{tab:performance-transposed} compares DICA with representative baselines on the POPE benchmark. DICA achieves the best performance in nearly all evaluation settings and datasets.  
On LLaVA-v1.5 under the AOKVQA/Popular setting, DICA reaches an F1-score of 85.99, outperforming all baselines. The concurrent improvements in accuracy and F1-score indicate that DICA effectively reduces false positive predictions, making the model more conservative in its production of affirmative answers. This directly alleviates object hallucination and improves fine-grained object understanding.

\noindent\textbf{Results on CHAIR.}
Table~\ref{tab:chair_results_modified} shows that DICA substantially reduces hallucinations in generative tasks. On both LLaVA-v1.5 and Qwen-VL, DICA achieves the lowest sentence-level and instance-level CHAIR scores among all methods.  
On LLaVA-v1.5, DICA reduces CHAIR\(_s\) to 41.8 and CHAIR\(_i\) to 11.5 while maintaining high recall, indicating that hallucination suppression does not come at the cost of descriptive coverage. On Qwen-VL, DICA attains similarly low hallucination rates with short outputs and preserves reasonable recall. DICA exhibits consistent hallucination suppression across architectures, demonstrating strong robustness.

\noindent\textbf{Results on MME.}  
As reported in Table~\ref{tab:mme}, DICA achieves the highest overall scores across all settings. On LLaVA-v1.5, DICA attains a total score of 661.5, exceeding all baseline methods. It performs best on the all subset and maintains leading performance on Counting and Position, indicating reliable perception of object presence and attributes.  
On Qwen-VL, DICA again achieves the highest total score of 648.3, with clear gains on the Existence, Position, and Color subsets. This suggests that DICA benefits from stronger visual encoders and improves cross-modal alignment. While performance on the Count subset is slightly lower than the best baseline, DICA remains the most stable method overall.

\begin{table}[t]
    \centering
    \small
    \renewcommand{\arraystretch}{1.0}
    \setlength{\tabcolsep}{2pt}
    \begin{tabular}{llccccc}
    \hline
     & Method & $\text{CHAIR}_{s} \downarrow$ & $\text{CHAIR}_{i} \downarrow$ & $\text{Recall} \uparrow$ & Length \\
    \hline
    \multirow{5}{*}{\rotatebox[origin=c]{90}{LLaVA-v1.5}} & sampling & 52.8 & 15.9 & 77.3 & 93.4 \\
     & VCD & 51.0 & 14.9 & 77.2 & 101.9 \\
     & M3iD & 56.2 & 17.0 & 79.3 & 97.1 \\
     & AvisC & 44.0 & 13.7 & 72.9 & 89.8 \\
     & MoD & \underline{42.6} & \underline{12.4} & 78.9 & 97.6 \\
     &DICA (ours)&\textbf{41.8}&\textbf{11.5}&77.8&95.8\\
    \hline
    \multirow{5}{*}{\rotatebox[origin=c]{90}{Qwen-VL}} & sampling & 2.8 & 3.0 & 31.0 & 5.3 \\
     & VCD & 1.4 & 1.2 & 30.8 & 4.0 \\
     & M3iD & 1.7 & 1.3 & 31.8 & 3.4 \\
     & AvisC & 1.6 & 1.6 & 32.0 & 4.4 \\
         & MoD  & \underline{0.8} & \underline{1.0} & 32.1 & 3.8 \\
     &DICA (ours)& \textbf{0.8}&\textbf{1.0} &27.0 &5.7 \\
    \hline
    \end{tabular}
    \caption{Results on CHAIR benchmark. Lower CHAIR$_{s}$ and CHAIR$_{i}$, along with higher Recall, correspond to better performance.}
    \label{tab:chair_results_modified}
\end{table}

\subsection{Ablation Studies}
To verify the effectiveness of the two key components in DICA, we conduct ablation studies on the POPE benchmark. As shown in Table~\ref{tab:performance-transposed}, removing either module results in a noticeable performance drop, confirming that both VAE-GDC and OIC-GGR are indispensable. The degradation is more pronounced when VAE-GDC is removed. This is primarily because attention drift occurs more frequently during inference, as indicated by the trigger statistics in Figure~\ref{fig:frequency}. Moreover, VAE-GDC operates at the attention level, directly reinforcing the visual grounding foundation, while OIC-GGR intervenes later at the output distribution level. Consequently, VAE-GDC serves a more foundational role in suppressing hallucinations. When both modules are jointly enabled, their complementary effects result in the best overall performance, underscoring the importance of combining attention-level diagnostics with output-level guidance.

\subsection{Analysis and Discussion}

\begin{figure}[t]
    \centering

    \includegraphics[width=1.0\linewidth]{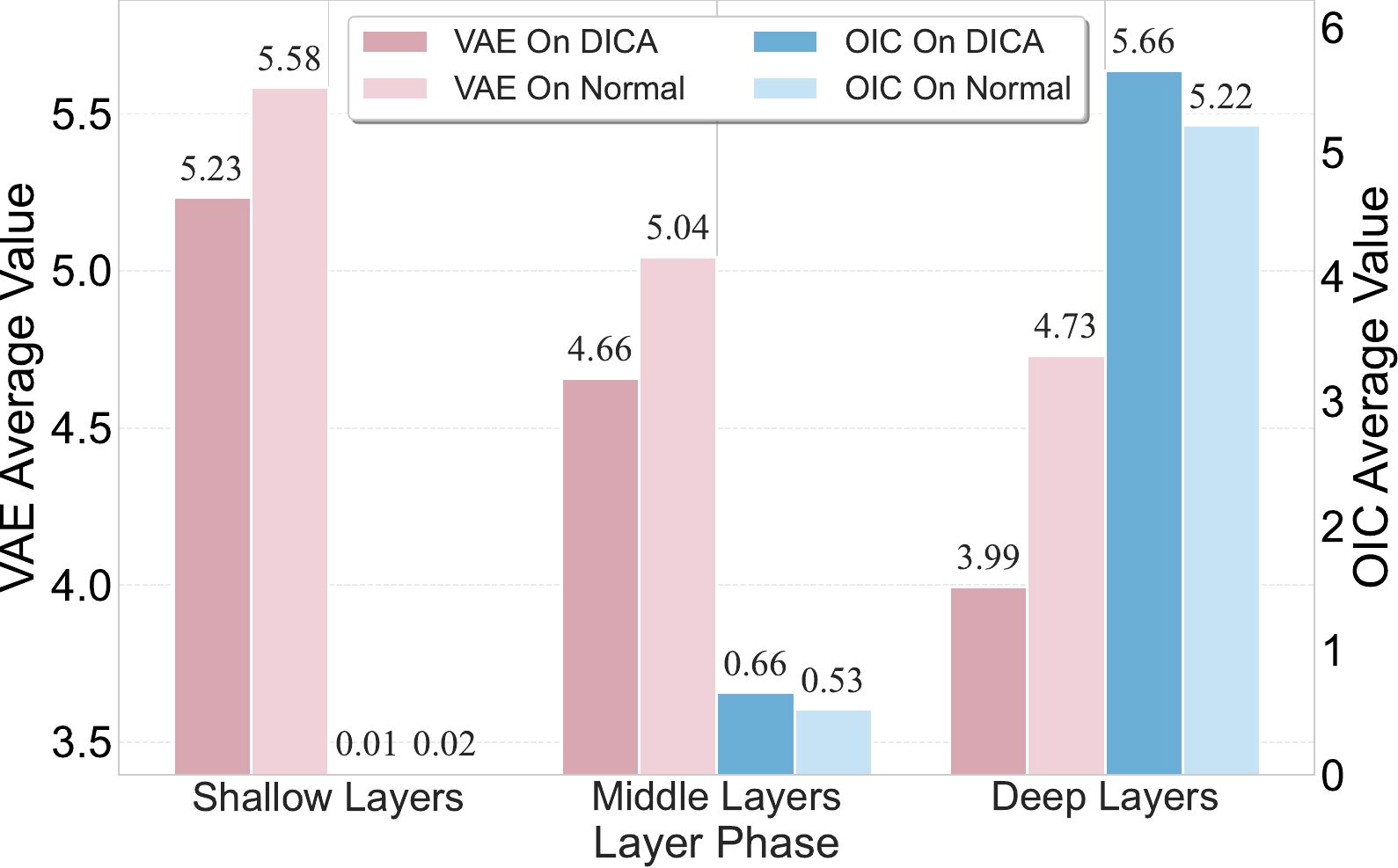}
    \caption{The impact of DICA on VAE and OIC within POPE.}
    \label{fig:vae and oic change}
\end{figure}

\textbf{The Mean Variations of OIC and VAE Before and After DICA.} Figure \ref{fig:vae and oic change} presents the average changes in VAE and OIC across shallow, middle, and deep layers before and after applying DICA. Overall, VAE consistently decreases, indicating more focused visual attention and reduced attention drift, while OIC increases across all layers, reflecting stronger reliance on image cues. These trends demonstrate that DICA effectively enhances visual grounding and improves the stability of the model's reasoning process.

\begin{table}[t]
    \centering
    \small
    \renewcommand{\arraystretch}{1.0}
    \setlength{\tabcolsep}{2pt}
    \begin{tabular}{llcccccc}
        \toprule
        & Method & \multicolumn{2}{c}{Object-level} & \multicolumn{2}{c}{Attribute-level} & \multirow{2}{*}{Total} \\
        \cmidrule(lr){3-4} \cmidrule(lr){5-6}
         & & Existence & Count & Position & Color & \\
        \midrule
        \multirow{5}{*}{\rotatebox[origin=c]{90}{LLaVA-v1.5}} & sampling & 170.0 & 103.3 & 108.3 & 128.3 & 510.0 \\
         & VCD & 180.0 & 110.0 & 108.3 & 133.3 & 531.7 \\
         & M3ID & 185.0 & 118.3 & 121.7 & 128.3 & 553.3 \\
         & AvisC & 195.0 & 116.7 & 131.7 & 153.3 & 596.7 \\
         & MoD  & 195.0 & 141.7 & 126.7 & 175.0 & 638.3 \\
         &DICA (ours) &\textbf{195.0}&\textbf{148.3} & \textbf{143.2}&\textbf{175.0} &\textbf{661.5} \\
        \midrule
        \multirow{5}{*}{\rotatebox[origin=c]{90}{Qwen-VL}} & sampling & 160.0 & 143.3 & 113.3 & 165.0 & 581.7 \\
         & VCD & 165.0 & 140.0 & 113.3 & 175.0 & 593.3 \\
         & M3ID & 165.0 & 143.3 & 103.3 & 175.0 & 586.7 \\
         & AvisC & 160.0 & 145.0 & 113.3 & 160.0 & 578.3 \\
         & MoD  & 170.0 & \textbf{160.0} & 103.3 & 180.0 & 613.3 \\
       &DICA (ours) &\textbf{170.0} &155.0 &\textbf{133.3} & \textbf{190.0}& \textbf{648.3}\\
        \bottomrule
    \end{tabular}
    \caption{Results on MME benchmark. The performance is measured by MME Score. The \textit{Total} column represents the sum of four individual results in each row.}
    \label{tab:mme}
\end{table}

\begin{figure}[t]
    \centering

    \includegraphics[width=1.0\linewidth]{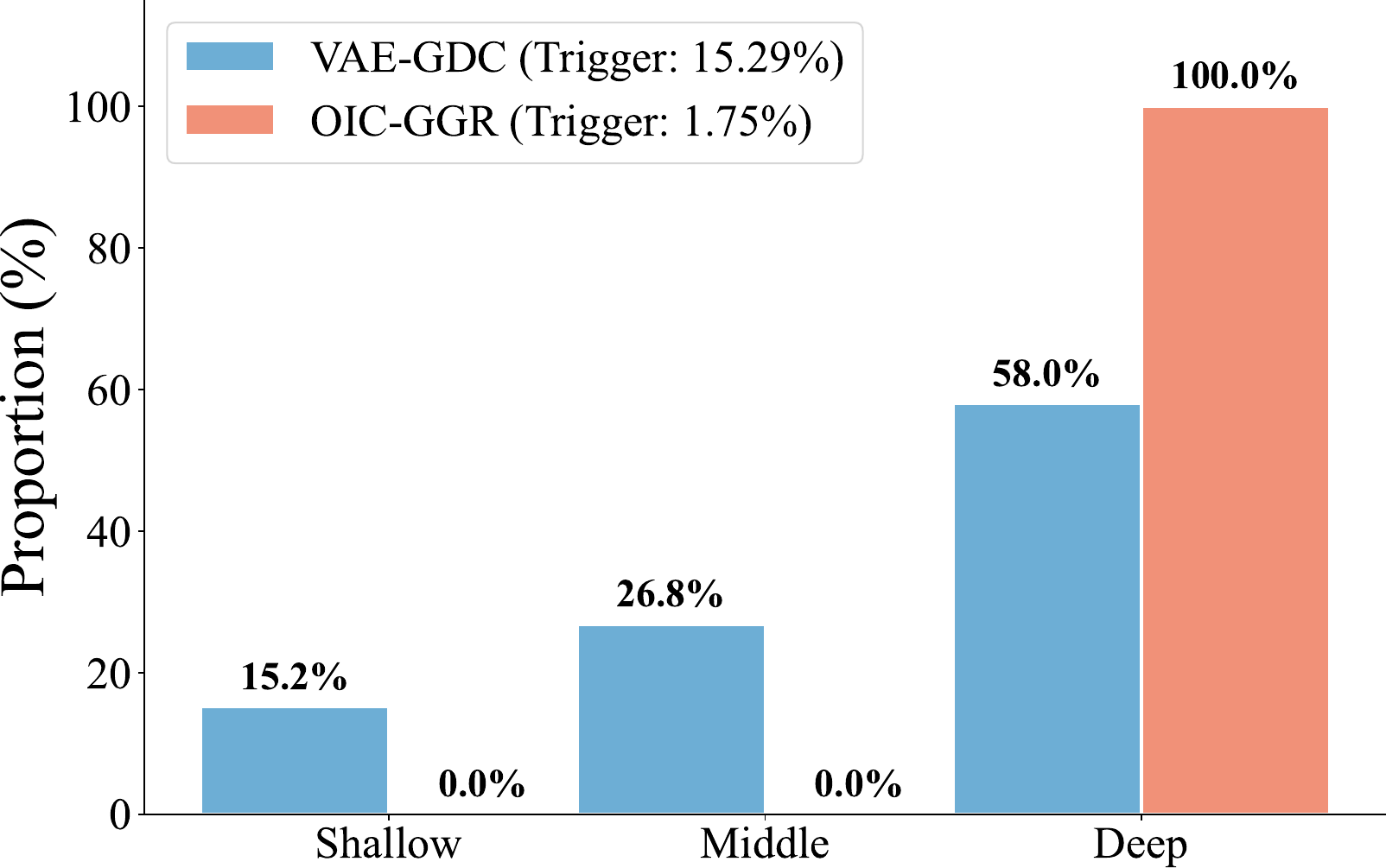}
    \caption{Trigger frequency and depth distribution of VAE-GDC and OIC-GGR on POPE.}
    \label{fig:frequency}
\end{figure}
\begin{table}[t!]
\centering
\small
\setlength{\tabcolsep}{4pt}
\begin{tabular}{lcccc}
\toprule
Repair outcome & W$\to$W & W$\to$R & R$\to$W & R$\to$R \\
\midrule
Overall (\%) & 8.2 & 4.9 & 1.3 & 86.9 \\
Conditional (\%) & 62.6 & 37.4 & 1.5 & 98.5 \\
\bottomrule
\end{tabular}
\caption{Repair success distribution on POPE. Here, W denotes wrong and R denotes right.}
\label{tab:repair_dist}
\end{table}

\noindent \textbf{Trigger Ratios and Triggered Layer Distribution of the Dual Module.}
Figure~\ref{fig:frequency} shows that VAE-GDC is triggered more frequently than OIC-GGR, and both modules are primarily activated in deeper layers. This indicates that attention drift occurs more often than underuse of visual evidence, with both failure modes emerging mainly in later reasoning stages. The higher trigger frequency of VAE-GDC suggests that maintaining focused visual attention is more challenging for MLLMs than preserving general image dependence, especially as decoding progresses.  

\noindent \textbf{Triggering Reliability and Correctness Transition.}
Table~\ref{tab:repair_dist}, DICA achieves highly precise interventions, with an R$\to$W rate of 1.3\%, indicating hardly any degradation of correct predictions. At the same time, it corrects 37.4\% of erroneous cases (W$\to$R), confirming that VAE and OIC reliably identify hallucination risks and guide the model back to relevant visual evidence via contrastive alignment. DICA operates as an on-demand intervention mechanism that strikes a balance between robustness and hallucination mitigation.

\section{Conclusion }
This study identifies two prevalent failure modes in MLLM-based visual question answering: attention drift and underuse of visual evidence. We propose the DICA framework, which leverages two information-theoretic indicators, VAE and OIC, to dynamically monitor model behavior. During normal reasoning, VAE decreases and OIC remains high; abnormal VAE surges or OIC drops trigger hallucinations, which DICA corrects via VAE-GDC and OIC-GGR, respectively. Extensive experiments across multiple models and benchmarks demonstrate that DICA  improves accuracy, effectively mitigates hallucinations, and validates the effectiveness of the proposed dual-indicator guided alignment.

\section*{Limitations}


There are three main limitations to our work:
\begin{itemize}
    \item Like other decoding-time hallucination mitigation methods, DICA introduces moderate computational overhead to enable real-time monitoring of visual grounding. This leads to a modest increase in latency compared to standard greedy decoding. 
    \item  The method relies on fixed threshold parameters, which may require manual adjustment for different models or downstream tasks, limiting its generalization across diverse MLLM architectures.
    \item DICA cannot compensate for perceptual blind spots inherent in the pretrained vision encoder or language model, and its performance is ultimately bounded by the base model's visual and linguistic capacities.
    
\end{itemize}

\section*{Acknowledgments}
This work was supported in part by the National Natural Science Foundation of China (NSFC) under Grant Nos. 61966038 and 62266051, and the Postgraduate Research and Innovation Foundation of Yunnan University
under Grant No.KC-252513133. The authors would like to thank the anonymous reviewers for their constructive comments.

\bibliography{custom}

\clearpage

\appendix

\section{Proof of Metric Applicability}

    
\label{sec:appendixz}
\label{sec:appendixa}
\
To validate that our proposed metrics, Visual Attention Entropy and Output-Image Correlation, faithfully reflect the human-like cognitive process of \emph{Global Exploration $\to$ Focal Exploitation}, we provide the following theoretical justifications based on information theory.

\noindent\textbf{1. Entropy as an Indicator of Visual Exploration Scope}

\noindent\textbf{Claim.} \textit{The Visual Attention Entropy $H^{(l,t)}$ serves as an information-theoretic proxy for the spatial scope of visual processing, characterizing the transition from global browsing to focal attention.}

\noindent\textit{Proof.} Let $\mathcal{V} = \{1, \dots, M\}$ be the set of visual token indices, and the attention distribution be $\mathbf{p} \in \Delta^{M-1}$. Consider two extreme cognitive states:
\begin{itemize}
  \item \textbf{Global Browsing (Exploration):} \(p_i = 1/M\) for all \(i\), entropy is maximized:
  \begin{equation}\label{eq:vae_exploration}
    \mathrm{VAE}(p) = - \sum_{i=1}^M \frac{1}{M} \log \frac{1}{M} = \log M.
  \end{equation}

  \item \textbf{Focal Attention (Exploitation):} \(p_k = 1\) for some token \(k\), others 0, entropy is minimized:
  \begin{equation}\label{eq:vae_exploitation}
    \mathrm{VAE}(p) = 0.
  \end{equation}
  \end{itemize}

Since Shannon entropy~\cite{6773024}  is concave and uniquely maximized by the uniform distribution, a decreasing sequence $\{H^{(l,t)}\}_t$ indicates a reduction in the effective receptive field, i.e., cognitive narrowing from scene-level to object-level focus. 

\vspace{1em}
\noindent\textbf{2. OIC as a Measure of Visual Evidence Contribution}

\noindent\textbf{Claim.} \textit{The OIC quantifies the information gain from visual tokens over the textual prior, measuring reliance on visual evidence.}

\noindent\textit{Proof.} Recall
\begin{equation}
\text{OIC}^{(l,t)} = \sum_v p_v \log \frac{p_v}{p_{\text{wi}, v}},
\end{equation}
which is the Kullback-Leibler divergence~\cite{kullback1951information}
\(
D_{\text{KL}}(P(\cdot|\text{Text, Image}) \parallel P(\cdot|\text{Text})).
\)
From a Bayesian view, text is the prior, image is the observation:
\begin{equation}
    \text{OIC}^{(l,t)} = \mathbb{E}_{v \sim P} \Big[ \log \frac{P(v \mid \text{Text, Image})}{P(v \mid \text{Text})} \Big].
\end{equation}
\begin{itemize}
    \item If visual info is redundant or ignored: $\text{OIC} \to 0$.
    \item If visual info provides critical evidence: $\text{OIC} \gg 0$.
\end{itemize}
Hence, OIC directly measures the informational contribution of visual perception to the decision-making process.

\vspace{1em}
\noindent\textbf{3. Modeling the Cognitive Evolution}

Based on the above analysis, the human VQA process can be described as a trajectory in $(\mathrm{VAE}, \text{OIC})$ space:

\[
\text{Process}:
\begin{cases}
\text{Phase I :} & t \text{ small}, \mathrm{VAE}^{(t)} \approx \log M, \\
                           & \text{OIC}^{(t)} \text{ moderate}, \\
\text{Phase II:} & t \uparrow,\mathrm{VAE}^{(t)} \downarrow, \\
                             & \text{context accumulates}, \\
\text{Phase III:} & t = t_{\text{ans}},\mathrm{VAE}^{(t)} \to 0, \\
                              & \text{OIC}^{(t)} \text{ maximized}.
\end{cases}
\]
This trajectory describes an idealized and interpretable pattern of cognitive evolution rather than a deterministic cognitive law.

\section{Proof of Metric Validity}
\label{sec:appendixb}


To understand the relationship between VAE and OIC, and how they reflect human-like cognitive processes in VQA, we analyze their theoretical connection through the lens of Information Theory and Transformer dynamics.

\paragraph{OIC as Information Gain.}
First, we observe that the definition of $\text{OIC}^{(l,t)}$ is equivalent to the KL divergence between the image-conditioned output distribution $\mathbf{p}^{(l,t)}$ and the text-only prior $\mathbf{p}^{(l,t)}_{\text{wi}}$:
\begin{equation}
    \text{OIC}^{(l,t)} = D_{\text{KL}}(\mathbf{p}^{(l,t)} \parallel \mathbf{p}^{(l,t)}_{\text{wi}}).
\end{equation}
This quantity measures the \textit{information gain} provided by the visual context at step $t$. A high OIC implies that the visual features significantly alter the model's prediction away from the pure language prior.

\paragraph{Attention Entropy as Feature Selectivity.}
The visual context vector $\mathbf{c}^{(l,t)}$ entering the feed-forward network is a weighted sum of visual tokens $V = \{\mathbf{v}_i\}$:
\begin{equation}
    \mathbf{c}^{(l,t)} = \sum_{i: m_i=1} p_i^{(l,t)} \mathbf{v}_i.
\end{equation}
We analyze two limiting cases of the attention entropy $H^{(l,t)}$ to demonstrate the evolution process:

\noindent\textbf{Phase 1: High Entropy (Global Exploration).}
During the initial stages of reasoning high $\mathrm{VAE}^{(l,t)}$, the attention distribution is dispersed, approximating a uniform distribution where $p_i^{(l,t)} \approx 1/N_v$ where $N_v$ is the number of visual tokens. The context vector becomes an average of the visual scene:
\begin{equation}
    \lim_{\mathrm{VAE} \to \mathrm{VAE}_{\text{max}}} \mathbf{c}^{(l,t)} \approx \frac{1}{N_v} \sum_{i} \mathbf{v}_i = \bar{\mathbf{v}}_{\text{global}},
\end{equation}
where $\bar{\mathbf{v}}_{\text{global}}$ represents generic scene semantics. While this informs the global context, it rarely conflicts strongly with the language before specific tokens, resulting in a \textit{moderate} divergence OIC. This mirrors the human cognitive phase of scanning for global semantics.

\noindent\textbf{Phase 2: Low Entropy (Focal Exploitation).}
As the model resolves the semantics low $\mathrm{VAE}^{(l,t)}$, the attention distribution becomes peaked and sparse. In the limit where $\mathrm{VAE}^{(l,t)} \to 0$, the probability mass concentrates on a single critical token $k$:
\begin{equation}
    \lim_{\mathrm{VAE} \to 0} \mathbf{c}^{(l,t)} = \mathbf{v}_k.
\end{equation}
In VQA, the visual token $\mathbf{v}_k$ typically contains specific evidence (e.g., color, count, text) that is orthogonal to the text context. The injection of this highly specific feature $\mathbf{v}_k$ causes the logits $\mathbf{z}^{(l,t)}$ to shift significantly from $\mathbf{z}^{(l,t)}_{\text{wi}}$, maximizing the KL divergence:
\begin{equation}
    \text{OIC}^{(l,t)} \propto \| \mathbf{z}^{(l,t)}(\mathbf{v}_k) - \mathbf{z}^{(l,t)}_{\text{wi}} \|. 
\end{equation}
This corresponds to the human cognitive phase of \textit{focusing on key details to formulate an answer}.

\paragraph{Conclusion.}
Therefore, the temporal evolution from high entropy to low entropy, accompanied by a rise in OIC, mathematically formalizes the cognitive strategy:
\begin{equation}
    \underbrace{\mathrm{VAE}^{(l,t)}_{\text{high}} \!\xrightarrow[]{\text{time}}\! \mathrm{VAE}^{(l,t)}_{\text{low}}}_{\text{Attention Focusing}} 
    \!\!\implies\!\! 
    \underbrace{\text{OIC}^{(l,t)}_{\text{low}} \!\xrightarrow[]{\text{time}} \!\text{OIC}^{(l,t)}_{\text{high}}}_{\text{Evidence Integration}}.
\end{equation}
This inverse correlation validates that the model accurately mimics the human VQA process, starting with global perception and narrowing down to the extraction of specific visual evidence.

\section{Related Work}
\label{Related Work}
\noindent \textbf{Multimodal Large Language Model.}
Inspired by the success of LLMs~\cite{touvron2023llama}, recent research has extended their capabilities to the multimodal domain, leading to the rapid development of MLLMs~\cite{liu2025attention,bai2023qwen}.  With the introduction of LLMs, MLLMs have increasingly adopted unified generative frameworks~\cite{liu2024improved,zhu2023minigpt}, which significantly enhance reasoning, generalization, and adaptability across diverse vision-language tasks. Typically, MLLMs are trained via a two-stage paradigm, including vision--language feature alignment pretraining followed by instruction-based fine-tuning~\cite{dai2023instructblip}. Recent studies have further incorporated human preference optimization techniques, such as RLHF~\cite{sun2024aligning} and preference fine-tuning~\cite{zhou2023analyzing}. Despite these advances, MLLMs still suffer from pervasive hallucination issues, where generated responses deviate from visual evidence, posing critical challenges to their reliability in real-world applications. This limitation underscores the need to develop more robust mechanisms to enhance visual grounding and factual consistency in MLLMs.

\noindent \textbf{Hallucination in MLLMs.}
Hallucinations, which generate content inconsistent with visual inputs or factual reality, have been widely studied in both language-only and vision-language settings \cite{yuan2026hybrid,yuan2025collaborative,ji2023survey,yao2023llm,liu2024exploring,li2023evaluating,yang2025evl,zheng2025enhanced,zheng2024enhancing,liu2025vision,11463006}. Due to the requirement of precise visual-textual alignment, hallucinations in MLLMs are often more severe, particularly in tasks such as visual question answering and image captioning, where models tend to over-rely on linguistic priors when visual grounding is unreliable or ambiguous \cite{biten2022let,wang2024mitigating,zhou2023analyzing}.

Existing hallucination mitigation methods in MLLMs can be broadly categorized into two classes according to whether they require additional training. 
\textbf{Training-based approaches} aim to suppress hallucinations by modifying model architectures or training objectives. Typical strategies include supervised fine-tuning on carefully curated anti-hallucination datasets, reinforcement learning with hallucination-aware rewards, or contrastive and counterfactual data construction that explicitly penalizes hallucinated outputs \cite{liu2023mitigating,cai2024vip,yu2024rlhf,villa2025eagle}. While effective in controlled settings, these methods are often costly in terms of data annotation and computation, and may compromise generalization when deployed beyond the training distribution.

In contrast, \textbf{training-free inference-time methods} attempt to regulate hallucinations without updating model parameters. A representative line of work leverages contrastive decoding, where hallucination-prone predictions are deliberately induced and contrasted against the original decoding stream to suppress spurious generations \cite{li2023contrastive,chuang2023dola}. In the multimodal domain, such hallucination-inducing signals are constructed through noisy or removed visual inputs \cite{leng2024mitigating,favero2024multi}, adversarial or negative prompts \cite{woo2025don,kim2023instructive}, or attention manipulation and decoding heuristics \cite{huang2024opera}. These approaches are attractive due to their flexibility and low deployment cost, and have shown promising empirical gains across benchmarks.

However, despite their differences, both training-based and training-free methods generally treat hallucination as a monolithic phenomenon and lack explicit mechanisms to diagnose its underlying causes. In particular, they fail to distinguish whether hallucinations arise from attention drift, or from underuse of visual evidence. As a result, existing methods tend to apply uniform correction strategies, such as globally suppressing language bias or enforcing stronger visual grounding, which may be misaligned with the actual failure mode and can even degrade correct predictions.

Motivated by this limitation, our proposed DICA departs from one-size-fits-all correction by explicitly diagnosing hallucination patterns during inference and performing targeted, adaptive interventions. By jointly monitoring VAE and OIC, DICA identifies whether hallucinations are caused by attention drift or underuse of visual evidence, and accordingly activates distinct mechanisms to recalibrate the reasoning process.

\section{More Eexperimental Details}
\label{sec:appendixd}
\subsection{Metrics}

\noindent\textbf{Metrics on MME. }In the MME benchmark, each image $i \in I$ is paired with two similar questions designed to elicit ``yes'' and ``no'' responses, denoted as $\{q_i^{y}, q_i^{n}\}$. Two metrics are used to evaluate model performance: the question-level accuracy (Acc) and the image-level accuracy (Acc$^+$).

The question-level accuracy (Acc) is defined as:
\begin{equation}
\begin{aligned}
\mathrm{Acc} =
&\;\frac{
\sum_{i \in I} \mathbb{I}\!\left[\mathrm{MLLMs}(i, q_i^{y}) = \mathrm{\text{"}Yes\text{"}}\right]
}{
|I|\times 2
}
\\[4pt]
&\;+
\frac{
\sum_{i \in I} \mathbb{I}\!\left[\mathrm{MLLMs}(i, q_i^{n}) = \mathrm{\text{"}No\text{"}}\right]
}{
|I|\times 2
}.
\end{aligned}
\end{equation}
The image-level accuracy (Acc$^+$) is defined as:
\begin{equation}
\begin{aligned}
\mathrm{Acc}^{+} =
&\;\frac{
\sum_{i \in I}
\mathbb{I}\!\left[
\mathrm{MLLMs}(i, q_i^{y}) = \mathrm{\text{"}Yes\text{"}}
\right]
}{
|I|
}
\\[4pt]
&\;\cdot\;
\frac{
\mathbb{I}\!\left[
\mathrm{MLLMs}(i, q_i^{n}) = \mathrm{\text{"}No\text{"}}
\right]
}{
1
}.
\end{aligned}
\end{equation}
The final MME Score is computed as:
\begin{equation}
\mathrm{MME\ Score} = \mathrm{Acc} + \mathrm{Acc}^{+}.
\end{equation}

\noindent\textbf{Metrics on CHAIR. }The CHAIR benchmark evaluates hallucinations in image captioning using two metrics: CHAIR$_i$ and CHAIR$_s$.

The instance-level hallucination rate is:
\begin{equation}
\mathrm{CHAIR}_i =
\frac{
|\{\text{hallucinated objects}\}|
}{
|\{\text{all mentioned objects}\}|
}.
\tag{12}
\end{equation}
The sentence-level hallucination rate is:
\begin{equation}
\mathrm{CHAIR}_s =
\frac{
|\{\text{sentences with hallucinations}\}|
}{
|\{\text{all sentences}\}|
}.
\tag{13}
\end{equation}
Recall measures caption completeness:
\begin{equation}
\mathrm{Recall} =
\frac{
|\{\text{accurately mentioned objects}\}|
}{
|\{\text{ground-truth objects}\}|
}.
\tag{14}
\end{equation}

\subsection{Results}
\subsubsection{Additional Results on CHAIR}
\begin{table}[t]
\centering
\setlength{\tabcolsep}{2pt}
\renewcommand{\arraystretch}{1.0}
\caption{CHAIR hallucination results on LLaVA-1.5. $C_S$ / $C_I$ (lower is better).}
\label{tab:llava_small_head}
\begin{tabular}{lcccc}
\hline
\small Method & \multicolumn{2}{c}{\small \textbf{LLaVA-1.5 (512)}} & \multicolumn{2}{c}{\small \textbf{LLaVA-1.5 (64)}} \\
\small        & \small $C_S$ & \small $C_I$ & \small $C_S$ & \small $C_I$ \\
\hline
Greedy      & 45.0 & 14.7 & 20.6 & 6.2 \\
Nucleus     & 48.8 & 14.2 & 26.2 & 8.5 \\
Beam Search & 48.8 & 13.9 & 18.8 & 5.9 \\
DoLA        & 47.8 & 13.8 & 20.4 & 6.3 \\
OPERA       & 44.6 & 12.8 & 14.2 & 5.2 \\
DICA       & \textbf{41.8} & \textbf{11.5} & \textbf{10.4}& \textbf{4.3} \\
\hline
\end{tabular}
\end{table}
Table~\ref{tab:llava_small_head} presents hallucination evaluation results on LLaVA-1.5 using the CHAIR benchmark, comparing various decoding strategies under two visual token settings: 512 (standard) and 64 (compressed). Lower CHAIR\textsubscript{S} and CHAIR\textsubscript{I} scores indicate fewer hallucinations. DICA consistently achieves the lowest hallucination rates among all methods, demonstrating its effectiveness in grounding model outputs in visual evidence. Notably, DICA maintains strong performance in both the 512-token and 64-token settings, indicating its robustness regardless of whether standard or compressed visual representations are used.

\begin{table*}[htbp]
\centering
\small

\label{tab:pope_results}
\begin{tabular}{lll ccccc ccccc}
\toprule
\textbf{Dataset} & \textbf{Setting} & \textbf{Method} & \multicolumn{4}{c}{\textbf{LLaVA-v1.5}} & \multicolumn{4}{c}{\textbf{Qwen-VL}} \\
& & & \textbf{Acc} & \textbf{Pre} & \textbf{Rec} & \textbf{F1} & \textbf{Acc} & \textbf{Pre} & \textbf{Rec} & \textbf{F1} \\
\midrule

\multirow{18}{*}{MS-COCO} 
& \multirow{6}{*}{random} 
    & sampling      & 83.8 & 82.4 & 86.1 & 84.2 & 84.9 & 96.0 & 72.9 & 82.9 \\
& & VCD           & 85.0 & 82.7 & 86.1 & 84.2 & 85.5 & 96.0 & 71.1 & 83.6 \\
& & M3ID          & 86.1 & 83.2 & 86.8 & 85.0 & 85.3 & 95.1 & 74.2 & 83.4 \\
& & AvisC         & 82.3 & 78.4 & 89.3 & 83.5 & 82.9 & 96.2 & 68.5 & 80.0 \\
& & MoD         & 89.2 & 90.0 & 88.2 & 89.1 & 86.0 & 97.0 & 74.3 & 84.1 \\
&& DICA (Ours) &\textbf{89.2}& 94.1 &83.7&88.8& \textbf{87.4} &96.9 &77.4 &\textbf{86.0}\\
\cmidrule{2-11}

\multirow{5}{*}{} 
& \multirow{6}{*}{popular} 
    & sampling      & 82.0 & 79.7 & 85.9 & 82.6 & 84.0 & 94.7 & 72.1 & 81.9 \\
& & VCD           & 82.1 & 78.5 & 88.3 & 83.2 & 84.9 & 94.5 & 74.9 & 83.6 \\
& & M3ID          & 82.8 & 80.1 & 88.5 & 84.1 & 84.2 & 94.1 & 73.8 & 82.7 \\
& & AvisC         & 78.2 & 72.7 & 90.3 & 80.5 & 82.8 & 95.5 & 68.9 & 80.1 \\
& & MoD     & 85.7 & 84.1 & 88.1 & 86.1 & 85.6 & 96.3 & 74.0 & 83.7 \\
&& DICA (Ours) &\textbf{87.7}& 90.6 &83.1&\textbf{86.7}& \textbf{87.5} &96.7 &77.6 &\textbf{86.1}\\
\cmidrule{2-11}

\multirow{5}{*}{} 
& \multirow{6}{*}{adversarial} 
    & sampling      & 75.8 & 71.3 & 86.3 & 78.1 & 82.1 & 90.0 & 72.3 & 80.2 \\
& & VCD           & 76.3 & 71.5 & 87.3 & 78.7 & 84.0 & 90.6 & 74.9 & 82.0 \\
& & M3ID          & 77.1 & 71.8 & 87.6 & 78.9 & 83.2 & 90.4 & 73.1 & 80.8 \\
& & AvisC         & 74.2 & 68.4 & 89.9 & 77.7 & 81.2 & 91.9 & 68.5 & 78.5 \\
& & MoD    & 79.7 & 75.4 & 88.2 & 81.3 & 84.0 & 92.4 & 74.2 & 82.3 \\
&& DICA (Ours) &\textbf{83.2}& 88.6 &76.2&\textbf{82.0}& \textbf{84.8} &93.1 &75.21 &\textbf{83.7}\\
\cmidrule{1-11}

\multirow{18}{*}{A-OKVQA} 
& \multirow{6}{*}{random} 
    & sampling      & 81.8 & 76.4 & 92.1 & 83.5 & 86.8 & 93.2 & 79.5 & 85.8 \\
& & VCD           & 81.2 & 75.2 & 93.0 & 83.2 & 87.4 & 92.9 & 81.1 & 86.6 \\
& & M3ID          & 82.9 & 76.8 & 94.1 & 84.6 & 87.1 & 92.4 & 80.3 & 85.9 \\
& & AvisC         & 79.1 & 71.9 & 95.5 & 82.1 & 84.7 & 93.0 & 74.9 & 83.0 \\
& & MoD     & 86.5 & 81.3 & 94.7 & 87.5 & 87.8 & 94.4 & 80.4 & 86.9 \\
&& DICA (Ours) &\textbf{89.2}& 87.9 &91.0&\textbf{89.4}& \textbf{89.5} &94.3 &84.3 &\textbf{88.9}\\
\cmidrule{2-11}

\multirow{5}{*}{} 
& \multirow{6}{*}{popular} 
    & sampling      & 75.3 & 69.1 & 91.5 & 78.7 & 85.6 & 90.6 & 79.5 & 84.7 \\
& & VCD           & 74.7 & 68.2 & 92.5 & 78.5 & 86.3 & 89.5 & 81.2 & 85.1 \\
& & M3ID          & 75.8 & 69.8 & 92.1 & 79.4 & 85.9 & 90.7 & 79.2 & 84.6 \\
& & AvisC         & 71.8 & 64.7 & 95.6 & 77.2 & 83.9 & 90.9 & 75.5 & 82.4 \\
& & MoD    & 79.5 & 72.6 & 94.5 &  82.2 & 86.5 & 91.6 & 80.3 & 85.6 \\
&& DICA (Ours) &\textbf{88.6}& 90.8 &81.7&\textbf{86.0}& \textbf{88.2} &93.1 &82.5 &\textbf{87.5}\\
\cmidrule{2-11}

\multirow{5}{*}{} 
& \multirow{6}{*}{adversarial} 
    & sampling      & 67.4 & 61.8 & 91.2 & 73.7 & 80.4 & 80.1 & 80.9 & 80.5 \\
& & VCD           & 68.1 & 61.9 & 93.8 & 74.6 & 80.7 & 80.1 & 81.6 & 80.8 \\
& & M3ID          & 68.3 & 62.1 & 93.4 & 74.6 & 80.5 & 80.7 & 80.2 & 80.4 \\
& & AvisC         & 64.4 & 58.8 & 96.1 & 73.0 & 78.1 & 80.1 & 74.7 & 77.3 \\
& & MoD     & 69.1 & 62.6 & 94.7 & 75.4 & 81.0 & 81.2 & 80.6 & 80.9 \\
&& DICA (Ours) &\textbf{77.8}& 75.7 &81.8&\textbf{78.6}& \textbf{81.3} &81.7 &80.8 &\textbf{81.3}\\
\cmidrule{1-11}

\multirow{18}{*}{GQA} 
& \multirow{6}{*}{random} 
    & sampling      & 81.6 & 75.6 & 93.2 & 83.5 & 81.3 & 88.8 & 71.5 & 79.2 \\
& & VCD           & 82.2 & 76.0 & 94.1 & 84.1 & 82.0 & 87.6 & 74.5 & 80.5 \\
& & M3ID          & 83.3 & 76.8 & 94.0 & 84.5 & 82.4 & 88.1 & 72.8 & 79.7 \\
& & AvisC         & 79.0 & 71.4 & 96.7 & 82.2 & 80.5 & 89.9 & 68.6 & 77.8 \\
& & MoD    & 86.2 & 80.8 & 95.1 & 87.4 & 83.8 & 90.3 & 75.7 & 82.3 \\
&& DICA (Ours) &\textbf{89.1}& 92.6 &85.1&\textbf{88.7}& \textbf{86.7} &94.5 &78.1 &\textbf{85.4}\\
\cmidrule{2-11}

\multirow{5}{*}{} 
& \multirow{6}{*}{popular} 
    & sampling      & 73.1 & 66.7 & 92.5 & 77.5 & 75.9 & 78.1 & 72.0 & 74.9 \\
& & VCD           & 71.5 & 64.7 & 94.5 & 76.8 & 75.9 & 76.6 & 74.7 & 75.6 \\
& & M3ID          & 72.3 & 64.9 & 94.8 & 77.1 & 76.8 & 78.9 & 75.2 & 77.0 \\
& & AvisC         & 67.4 & 60.9 & 97.1 & 74.8 & 74.2 & 77.9 & 67.5 & 72.3 \\
& & MoD     & 74.0 & 66.8 & 95.3 & 78.6 & 79.8 & 82.4 & 75.7 & 78.9 \\
&& DICA (Ours) &\textbf{81.9}& 79.4 &85.5&\textbf{82.5}& \textbf{82.1} &86.4 &77.6 &\textbf{81.7}\\
\cmidrule{2-11}

\multirow{5}{*}{} 
& \multirow{6}{*}{adversarial} 
    & sampling      & 68.0 & 62.0 & 93.4 & 74.5 & 75.5 & 77.8 & 71.2 & 74.4 \\
& & VCD           & 67.6 & 61.5 & 94.4 & 74.5 & 76.7 & 77.8 & 74.7 & 76.2 \\
& & M3ID          & 67.2 & 61.0 & 93.9 & 74.0 & 77.1 & 78.6 & 74.9 & 76.7 \\
& & AvisC         & 64.1 & 58.5 & 96.7 & 72.9 & 75.5 & 80.5 & 67.4 & 73.4 \\
& & MoD     & 68.7 & 62.2 & 95.3 & 75.3 & 78.9 & 81.0 & 75.4 & 78.1 \\
&& DICA (Ours) &\textbf{78.4}& 74.8 &85.9&\textbf{79.9}& \textbf{79.2} &82.8 &74.5 &\textbf{78.2}\\
\bottomrule
\end{tabular}
\caption{Detailed evaluation results on POPE benchmark (Acc, Pre, Rec, F1) for LLaVA-v1.5 and Qwen-VL.}
\label{tab:performance_final}
\end{table*}

\subsubsection{Additional Results on POPE}
Table \ref{tab:performance_final} reports the complete experimental results on the POPE benchmark. DICA consistently outperforms nearly all existing methods, including VCD, M3ID, and other training-free and training-based baselines. On LLaVA-1.5 under the MS-COCO Popular setting, DICA achieves 87.7\% accuracy, outperforming Sample (82.0\%), VCD (82.1\%), M3ID (82.8\%), AVisC (78.2\%), and MoD (85.7\%). On Qwen-VL, DICA exceeds all competitors across all nine accuracy metrics. Overall, DICA achieves the strongest performance, particularly under adversarial and popular conditions, demonstrating superior visual grounding and effective hallucination mitigation during inference.
\subsection{More Analysis and Discussion}
\subsubsection{VAE and OIC Performance Changes with DICA}
\begin{figure}[t]
    \centering
    \includegraphics[width=0.9\linewidth]{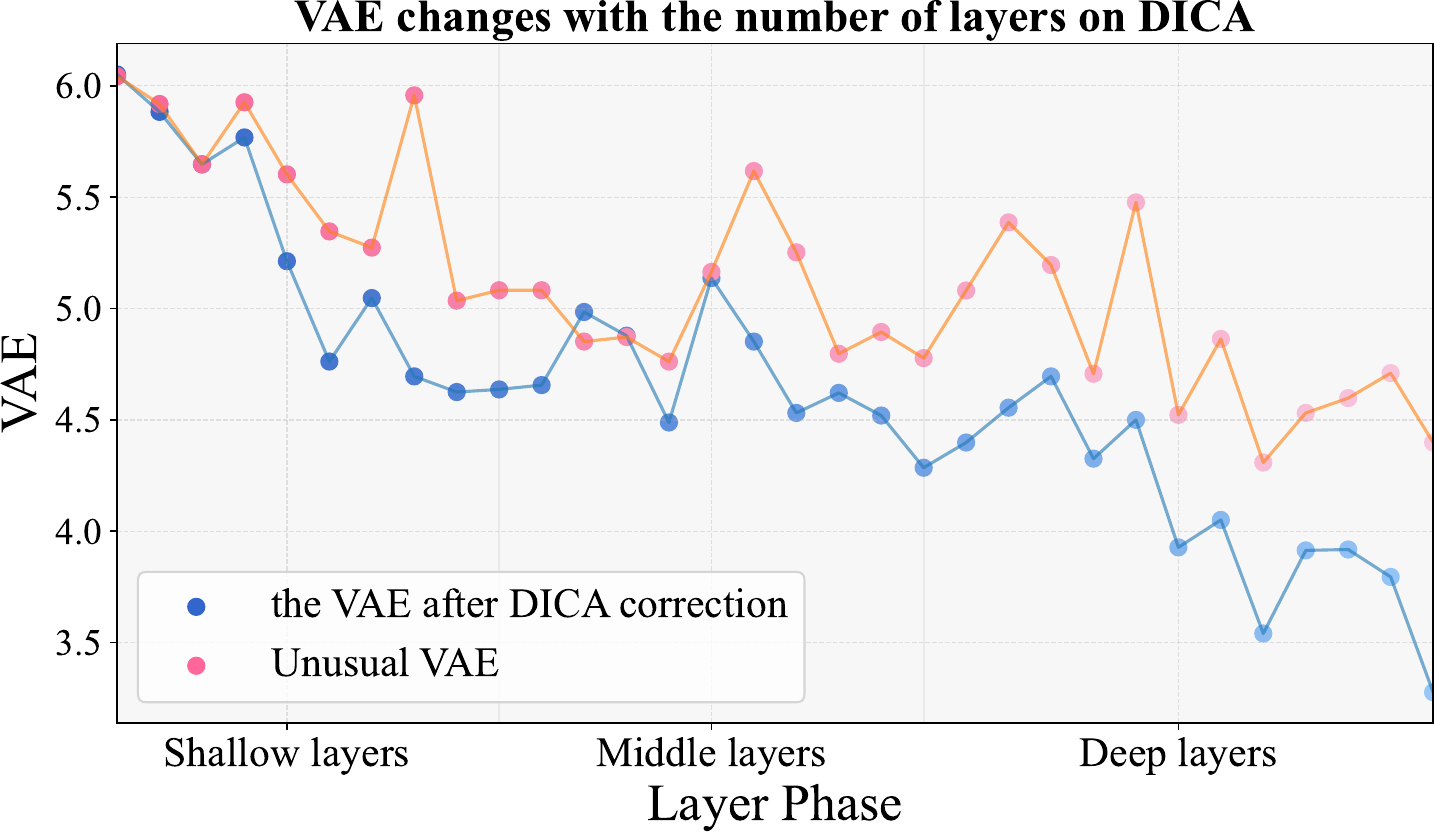}\\[2mm]
    \includegraphics[width=0.9\linewidth]{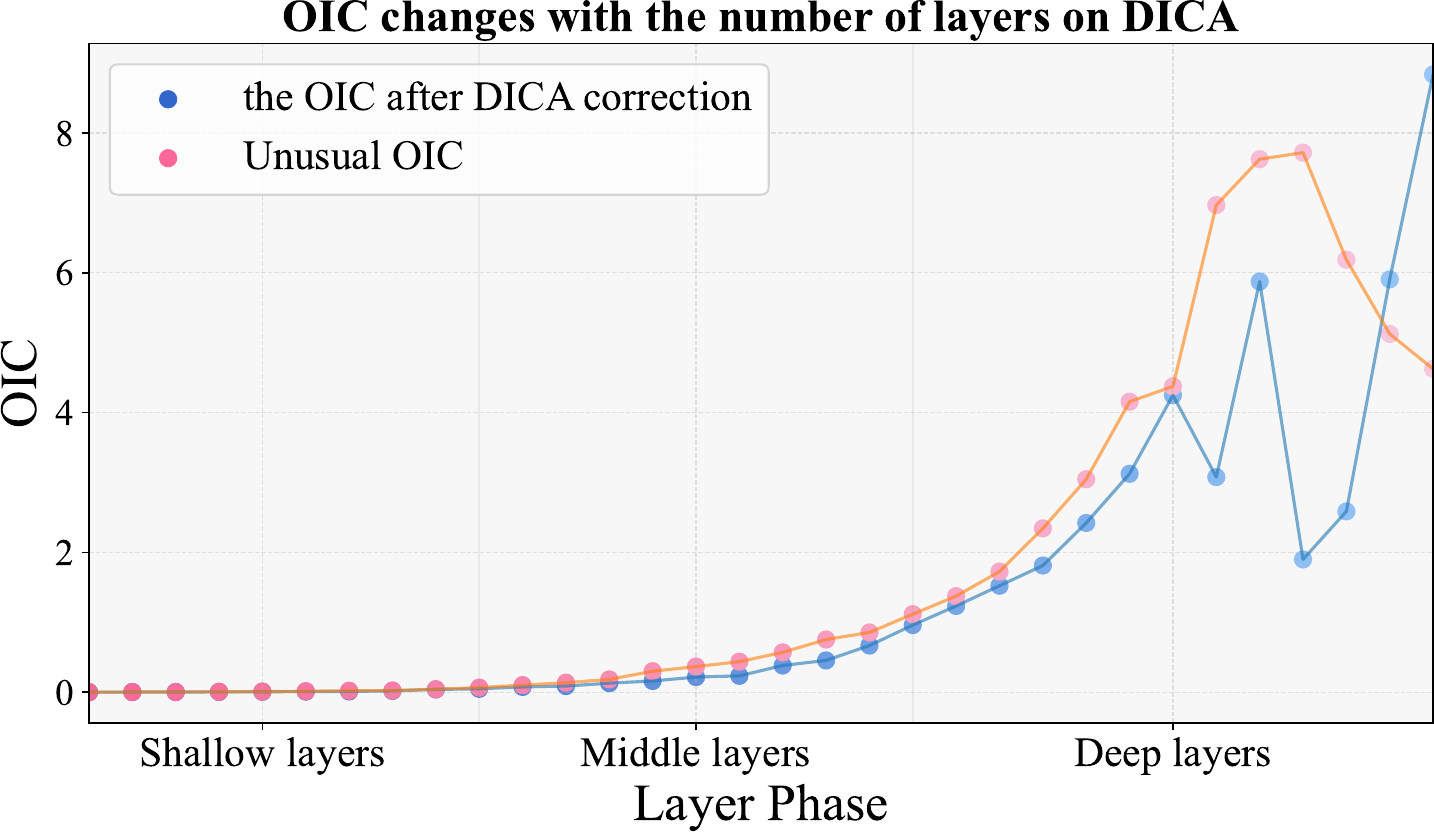}
    \caption{\textbf{Top:} VAE changes across layers phase on DICA. \textbf{Down:} OIC change across layers phase on DICA.}
    \label{fig:vae_and_oic_2}
\end{figure}
Figure~\ref{fig:vae_and_oic_2} illustrates the evolution of VAE and OIC across shallow, middle, and deep layers after applying the DICA intervention. The corrected trajectories demonstrate that VAE steadily decreases. Meanwhile, OIC consistently increases, reflecting a reasoning process that aligns with the human-like cognitive pattern of shifting from global exploration to focal exploitation. In contrast, the uncorrected anomalous steps reveal instances of attention drift, abnormally high VAE, and underuse of visual evidence, as well as abnormally low OIC. DICA effectively suppresses these anomalies, ensuring that visual attention remains focused and that model predictions stay grounded in image evidence throughout inference.

\subsubsection{Analysis of $\alpha_\mathrm{VAE}$ and $\alpha_\mathrm{OIC} $}
\label{Hyperparameter analysis}
\begin{figure}[t]
    \centering
    \includegraphics[width=1.0\linewidth]{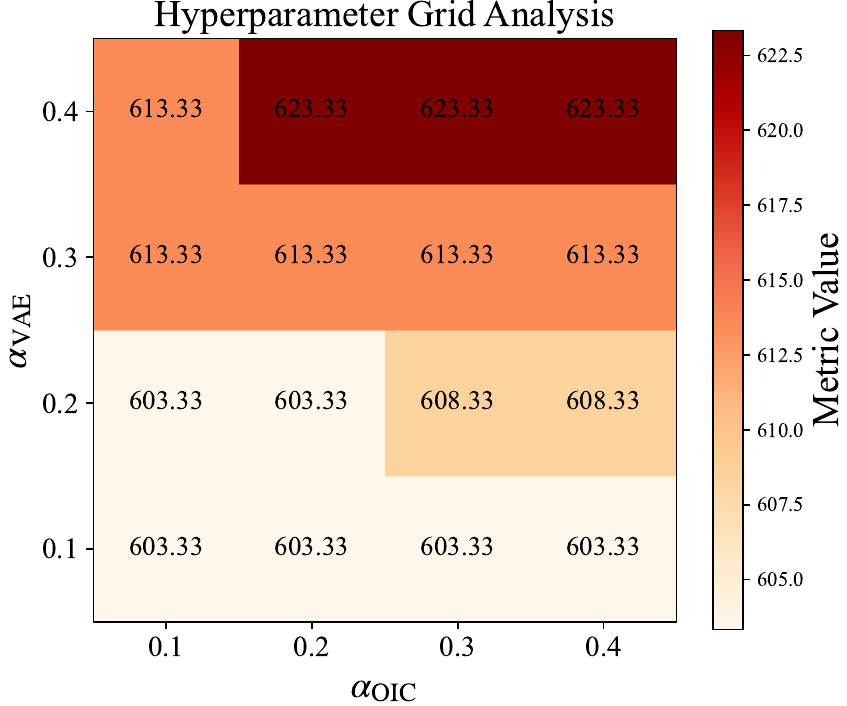}\\[2mm]
    \includegraphics[width=1.0\linewidth]{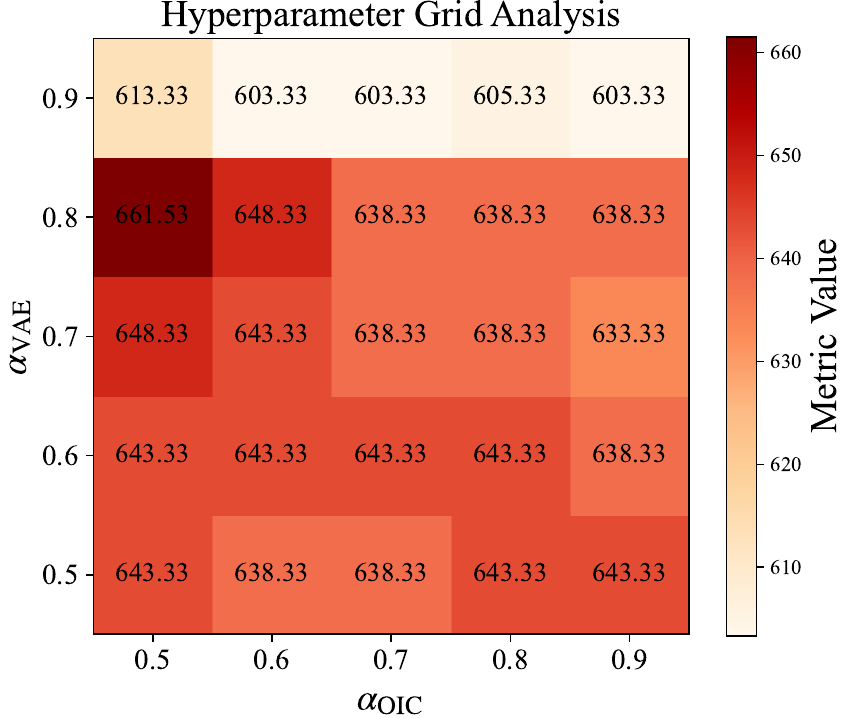}
    \caption{The upper figure corresponds to thresholds 0.1--0.4, and the lower figure to thresholds 0.5--0.9.}
    \label{fig:vae_and_oic_alpha}
\end{figure}
The sensitivity of DICA to the drift and grounding thresholds $\alpha_{\mathrm{VAE}}$ and $\alpha_{\mathrm{OIC}}$ is evaluated through a grid search on the MME benchmark. Figure~\ref{fig:vae_and_oic_alpha} top presents the results when both thresholds range from 0.1 to 0.4. In this low-threshold regime, performance remains at a relatively low level, with metric values concentrated between 603.33 and 608.33, showing no clear improvement.

When the thresholds are set below 0.5, DICA becomes overly sensitive, causing the intervention mechanisms to be frequently triggered. Normal fluctuations and minor noise in the reasoning process are often misidentified as abnormal signals, leading VAE-GDC and OIC-GGR to apply unnecessary corrections. Such over-correction disrupts otherwise stable inference trajectories and degrades overall performance.

In contrast, the moderate-to-high threshold regime shown in the down part of Figure~\ref{fig:vae_and_oic_alpha} effectively alleviates this issue. With $\alpha_{\mathrm{VAE}} = 0.8$ and $\alpha_{\mathrm{OIC}} = 0.5$, interventions are activated only when substantial attention drift or insufficient visual grounding occurs. This selective activation avoids noise-induced corrections while enabling targeted adjustments, resulting in the best performance of 661.53. These results indicate that slightly higher threshold values strike a better balance between intervention frequency and correction strength, which is crucial for DICA to achieve optimal effectiveness.
\subsubsection{Analysis of $\alpha_{d}$}
\begin{figure}[t]
    \centering
    \includegraphics[width=1.0\linewidth]{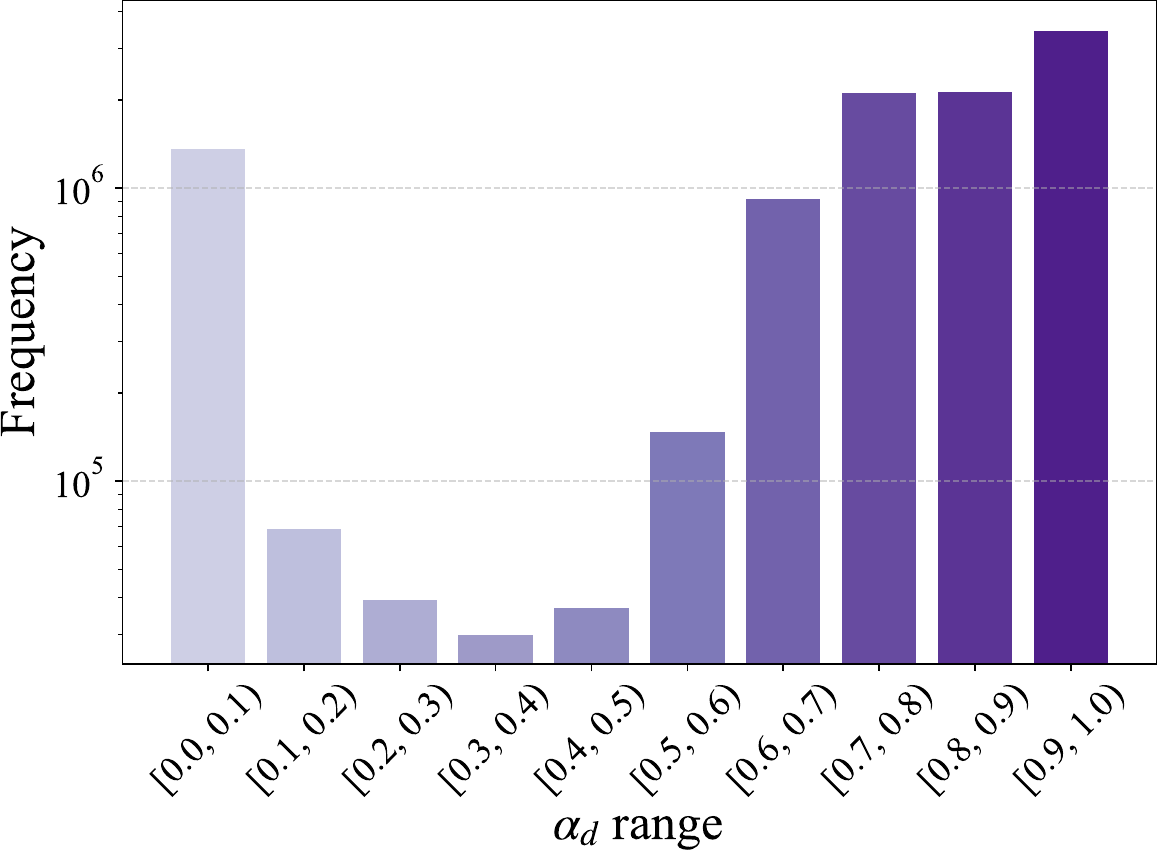}
    \caption{ The change of $\alpha_{d}$ On POPE.}
    \label{fig:alpha_t}
\end{figure}
\begin{figure}[t]
    \centering
    \includegraphics[width=1.0\linewidth]{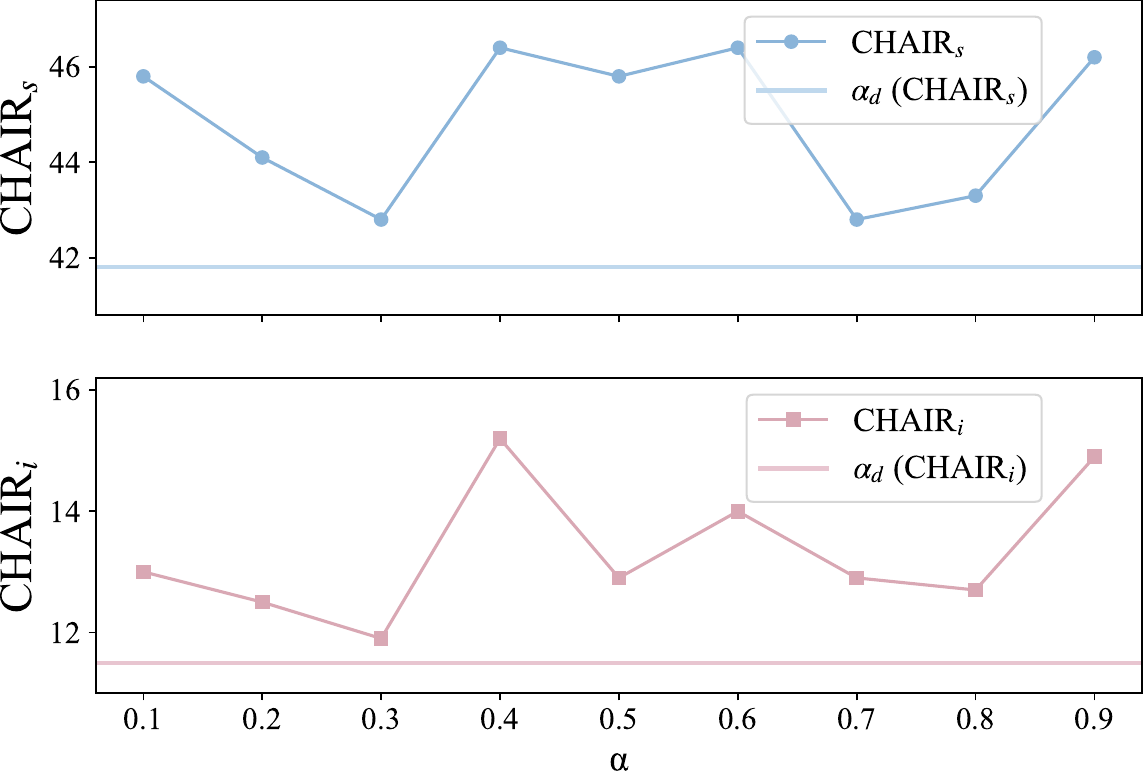}
    \caption{ Performance with fixed $\alpha$ on CHAIR.}
    \label{fig:alpha}
\end{figure}

The value of $\alpha_d$ exhibits a pronounced bimodal distribution during inference on the POPE benchmark Figure~\ref{fig:alpha_t}, with most values concentrated near 0 or 1. This shows that VAE-GDC's contrastive correction is selectively activated, either fully engaged or nearly inactive---rather than applied uniformly. Values in [0.0, 0.1) occur when the Normal and Descriptive paths produce highly divergent outputs, typically during stable, visually grounded reasoning. Here, the Descriptive path which generates open-ended descriptions naturally differs from the task-specific Normal path, and the low $\alpha_d$ prevents unnecessary correction, preserving fluency. Conversely, values in [0.9, 1.0) arise during attention drift: both paths yield diffuse, high-entropy distributions due to lost visual focus, making their outputs unusually similar. The near unity $\alpha_d$ triggers strong contrastive decoding to recover a grounded prediction. The scarcity of intermediate values confirms that DICA acts as a sparse, decisive intervention, activating only when clear anomalies are detected.

We further fix $\alpha_d$ to a constant value $\alpha$ and conduct experiments on the CHAIR dataset. As shown in Figure~\ref{fig:alpha}, our results reveal a clear relationship between the blending weight $\alpha$ and hallucination behavior. When $\alpha$ is close to 0, the model heavily relies on the Normal path, which is tightly grounded in visual input for the specific task. This leads to strong visual alignment and the lowest CHAIR scores, indicating minimal object hallucination. As $\alpha$ increases toward 1, the influence of the Descriptive path grows, introducing more open-ended and fluent language patterns but also increasing the risk of describing non-existent objects, as reflected by rising CHAIRs and CHAIRi. The best performance is consistently achieved at small $\alpha$ values, confirming that precise visual grounding enabled by prioritizing the Normal path is crucial for hallucination suppression in captioning. This empirical trend validates the design of the adaptive $\alpha_d$, which automatically favors the Normal path during stable reasoning and only invokes the Descriptive path when attention drift necessitates corrective intervention.

\subsubsection{Analysis of $\beta_d$}

\begin{figure}[t]
    \centering
    \includegraphics[width=1.0\linewidth]{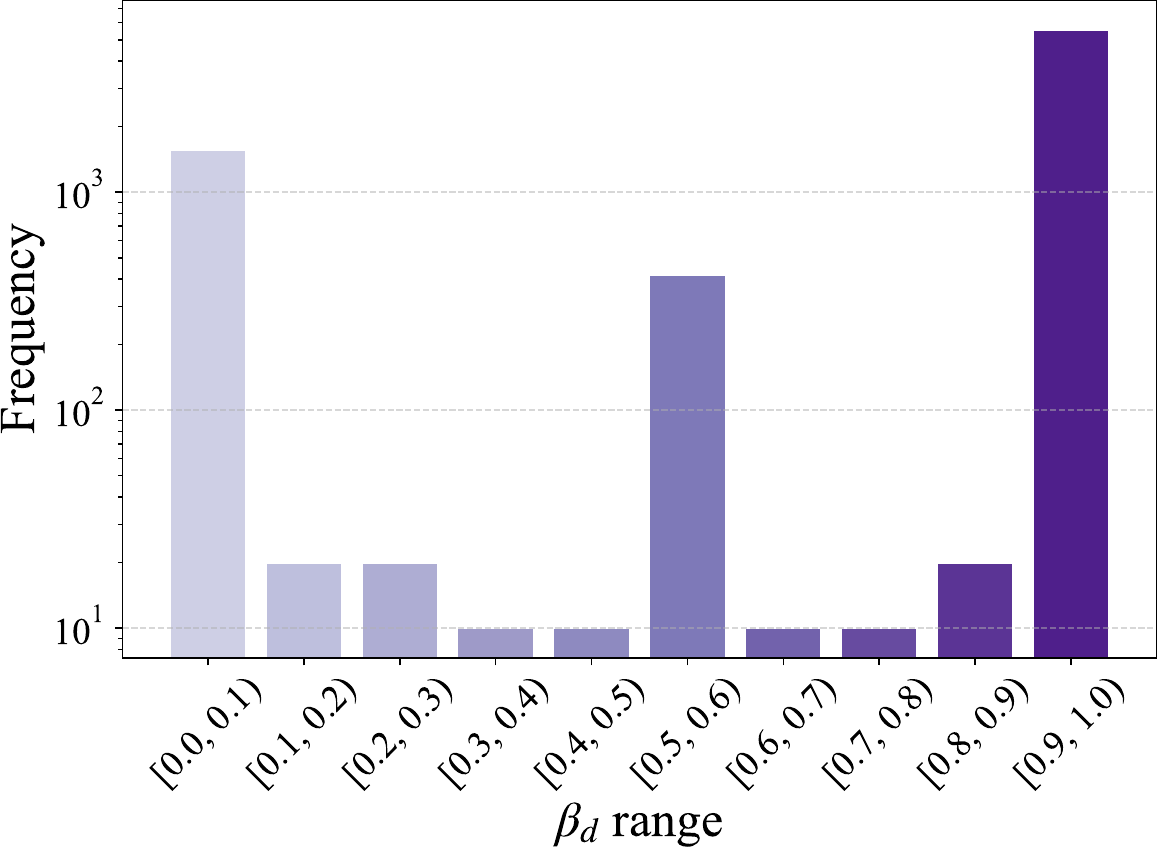}
    \caption{ The change of $\beta_{d}$ on POPE.}
    \label{fig:beta_t}
\end{figure}
The value of $\beta_d$ exhibits a pronounced bimodal distribution during inference on the POPE benchmark. As shown in Figure \ref{fig:beta_t}, the majority of $\beta_d$ values are clustered near the extremes of $0$ and $1$, rather than being uniformly distributed. Specifically, the high-density peak near $0$ corresponds to stable reasoning phases where the model maintains a high correlation between its output and the relevant visual evidence, indicating that the internal attention mechanism is functioning correctly without excessive intervention. Conversely, the sharp peak near $1$ signifies instances where the OIC drops significantly, reflecting potential attention drift or underuse of visual evidence. The fact that $\beta_d$ effectively polarizes these two states demonstrates that DICA can act as a precise binary-like trigger. This adaptive mechanism ensures that contrastive alignment is only aggressively applied when a deviation from visual evidence is detected, thereby suppressing hallucinations while preserving the linguistic fluency of the model in normal states.

\begin{figure}[t]
    \centering
    \includegraphics[width=1.0\linewidth]{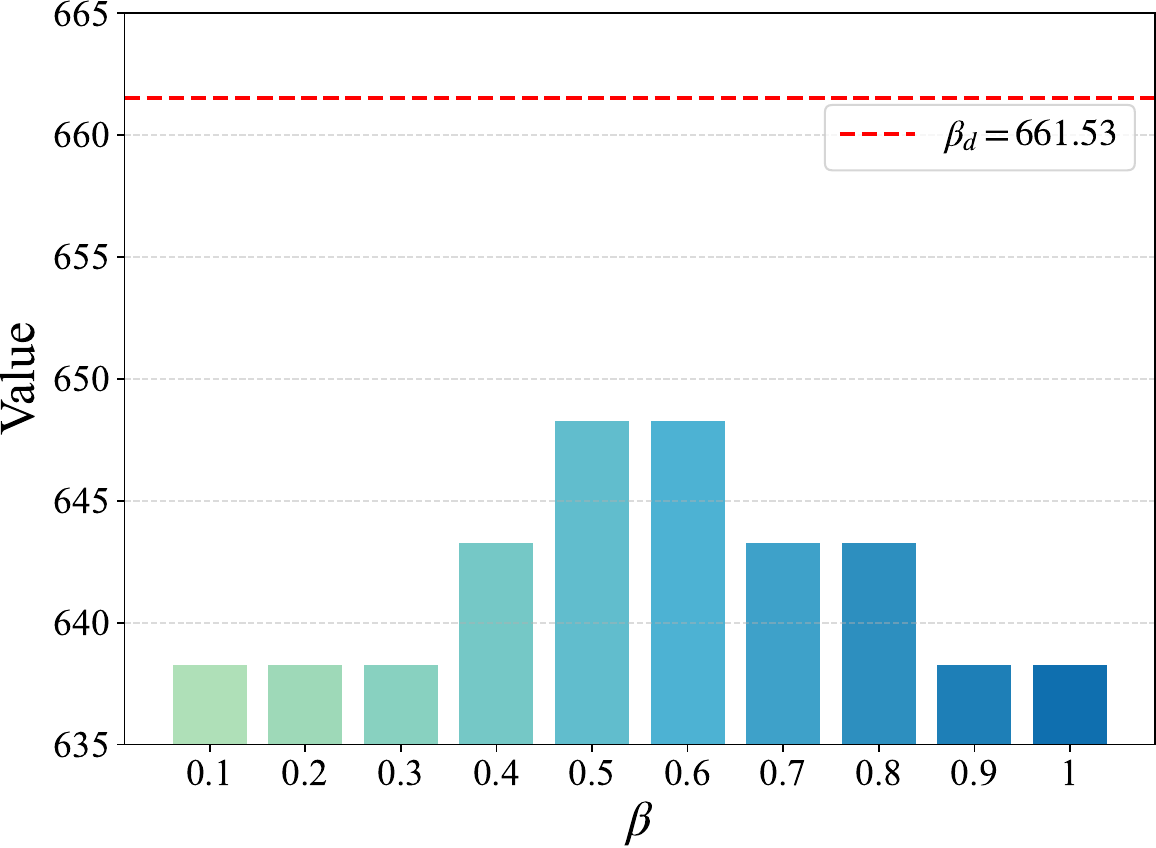}
    \caption{Performance with fixed $\beta$ on MME.}
    \label{fig:beta}
\end{figure}
We fixed the original~$\beta_d$~value from the paper and replaced it with a unified constant hyperparameter~$\beta$. As shown in Figure~\ref{fig:beta}, the evaluation on the MME benchmark shows that the original setting with~$\beta_d$~achieves a score of~661.53, which significantly outperforms all configurations using fixed~$\beta$~values. Specifically, across the tested range of~$\beta$~from~0.1~to~1.0, the best performance is only~648.33, which is clearly lower than the score obtained with~$\beta_d$. This result demonstrates that the dynamic design of~$\beta_d$~more effectively balances generation fluency and visual grounding, leading to superior overall performance and further validating the effectiveness of the OIC-GGR module in DICA.

\subsubsection{Effectiveness of DICA in Visible}
\label{Effectiveness of DICA in Visible}
\begin{figure*}[t]
    \centering

    \includegraphics[width=1.0\linewidth]{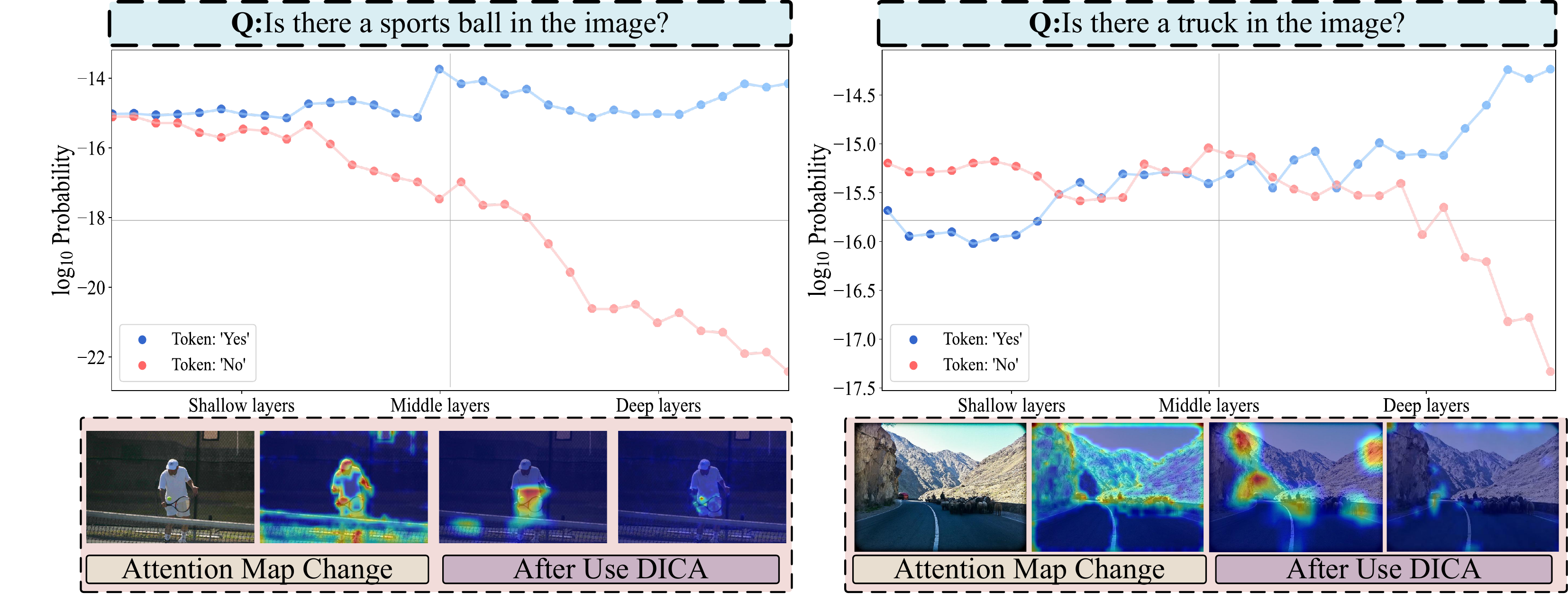}
    \caption{Evolution of answer log-probabilities and attention maps across layers applying DICA.}
    \label{fig:After_dica_results}
\end{figure*}
Without DICA, the model exhibits clear signs of \textbf{attention drift} and \textbf{underuse of visual evidence} as illustrated in Figure~\ref{fig:Adver_results}. For instance, when answering ``Is there a sports ball in the image?'', the attention map becomes diffused in deeper layers, shifting away from the relevant object region; simultaneously, the log-probability of the correct answer drops sharply, leading the model to output an incorrect ``No''. A similar failure occurs for the query ``Is there a truck in the image?'', where visual cues are largely ignored despite their presence, resulting in hallucinated responses.
`````
In contrast, after applying DICA see Figure~\ref{fig:After_dica_results}, our dual-indicator monitoring detects these anomalies in the middle-to-late decoding stages: a sudden rise in VAE triggers VAE-Guided Drift Correction. Meanwhile, a drop in OIC activates OIC-Guided Grounding Restoration. As a result, the model's attention is realigned to the semantically relevant regions, and the answer log-probabilities stabilize and converge toward the correct label. The final predictions become both visually grounded and accurate, demonstrating DICA's effectiveness in restoring faithful cross-modal reasoning.
\begin{table*}[t]
\centering
\setlength{\tabcolsep}{6pt}
\renewcommand{\arraystretch}{1.0}
\small
\begin{tabular}{lcccc}
\toprule
\textbf{Approach} &
\textbf{Training Time} &
\textbf{Data Prep Overhead} &
\textbf{Inference Cost} &
\textbf{Deployment Agility} \\
\midrule
Retraining / RLHF
& High
& High
& $1\times$
& Low \\

Data Engineering
& Medium (fine-tuning required)
& Very High
& $1\times$
& Low \\

Beam Search ($k=5$)
& Zero
& Zero
& $5\times$
& High \\

\textbf{DICA (Ours)}
& \textbf{Zero}
& \textbf{Zero}
& $\sim$2.5$\times$
& \textbf{Very High} \\
\bottomrule
\end{tabular}

\caption{
Comparison of time and deployment costs across hallucination mitigation paradigms.
Training Time is measured in GPU-hours.
Data preparation overhead refers to the qualitative human and computational effort required to curate or generate additional supervision signals before training or inference.
The inference cost is reported as a multiplicative factor relative to standard greedy decoding.
}
\label{tab:time-cost-comparison}
\end{table*}
\subsubsection{Efficiency Analysis}
\label{efficiency analysis}

We analyze the computational overhead introduced by DICA and compare it with the total time cost of alternative hallucination mitigation strategies, including training-based and data-engineering-based approaches. During inference, standard autoregressive decoding requires one forward pass per token. In contrast, DICA computes lightweight indicators from three computational streams: the normal path, the text-only path, and the descriptive path, which yields a theoretical upper bound of approximately three forward passes per token. However, the actual wall-clock latency increase is significantly smaller. This is because the vision encoder, the heaviest module in MLLMs, runs only once. Meanwhile, subsequent computations operate solely within the decoder layers and therefore benefit heavily from KV-cache reuse. Moreover, the additional streams are executed within the same decoding iteration. They are processed jointly on modern GPUs, whose parallel throughput leads to sub-linear latency growth relative to the number of streams. Another important observation is that DICA activates its computationally expensive correction modules only when the VAE or OIC indicators exceed their respective thresholds, meaning that the overhead is sparse rather than incurred at every decoding step. For example, on LLaVA-1.5, DICA adds roughly 0.04--0.08 seconds per token relative to greedy decoding. For typical VQA outputs under 20 tokens, the end-to-end delay remains below 1.5 seconds, which is well within the interactive latency tolerance.

The more substantial efficiency advantage of DICA appears when considering the full lifecycle cost of deploying a hallucination mitigation method. Unlike approaches requiring supervised fine-tuning, RLHF, or counterfactual data construction, DICA introduces zero training cost and zero data preparation cost. In comparison, beam search incurs a 3--5$\times$ inference multiplier. Table~\ref{tab:time-cost-comparison} summarizes the comparison across methods. Despite a moderate increase in inference-time computation (approximately 2.5$\times$), DICA eliminates the significantly higher cost associated with retraining or building large-scale, curated datasets.

\subsubsection{Performance on Qwen3}
To further verify the generalization ability of our method, we conducted supplementary experiments on a more recent backbone model, Qwen3~\cite{yang2025qwen3technicalreport}, under the POPE benchmark. Specifically, we compared the original Qwen3 baseline with Qwen3 enhanced by our DICA framework.

Table~\ref{tab:qwen3_pope} presents the detailed results. Across all datasets and evaluation settings, DICA consistently improves both Accuracy and F1 score over the original Qwen3 model. Notably, performance gains are observed not only under Random splits but also under more challenging Popular and Adversarial settings, which are designed to evaluate robustness against hallucination biases.

These results indicate that DICA is not tightly coupled with a specific architecture but instead acts as a general and transferable hallucination mitigation framework. 

\begin{table}[t]
\centering
\setlength{\tabcolsep}{4pt}
\renewcommand{\arraystretch}{1.0}
\small
\begin{tabular}{llcccc}
\hline
Dataset & Setting & \multicolumn{2}{c}{Qwen3(Base)} 
& \multicolumn{2}{c}{Qwen3(DICA)} \\
 &  & Acc & F1 & Acc & F1 \\
\hline

\multirow{3}{*}{MSCOCO}
& Random & 89.50 & 88.67 & 91.73 & 91.15 \\
& Popular & 87.47 & 86.73 & 89.13 & 88.68 \\
& Adversarial & 85.70 & 85.11 & 87.23 & 86.95 \\
\hline
\multirow{3}{*}{AOKVQA}
& Random & 92.80 & 92.71 & 94.10 & 94.07 \\
& Popular & 89.00 & 89.35 & 90.30 & 90.50 \\
& Adversarial & 82.70 & 84.14 & 84.37 & 85.52 \\
\hline
\multirow{3}{*}{GQA}
& Random & 91.70 & 91.55 & 92.77 & 92.59 \\
& Popular & 87.60 & 88.01 & 89.37 & 89.48 \\
& Adversarial & 83.87 & 84.77 & 85.00 & 85.80 \\

\hline
\end{tabular}
\caption{Performance comparison on the POPE benchmark. Base denotes the original Qwen3 model.}
\label{tab:qwen3_pope}
\end{table}
\section{More Case Studies}
\begin{figure*}[t]
    \centering

    \includegraphics[width=1.0\linewidth]{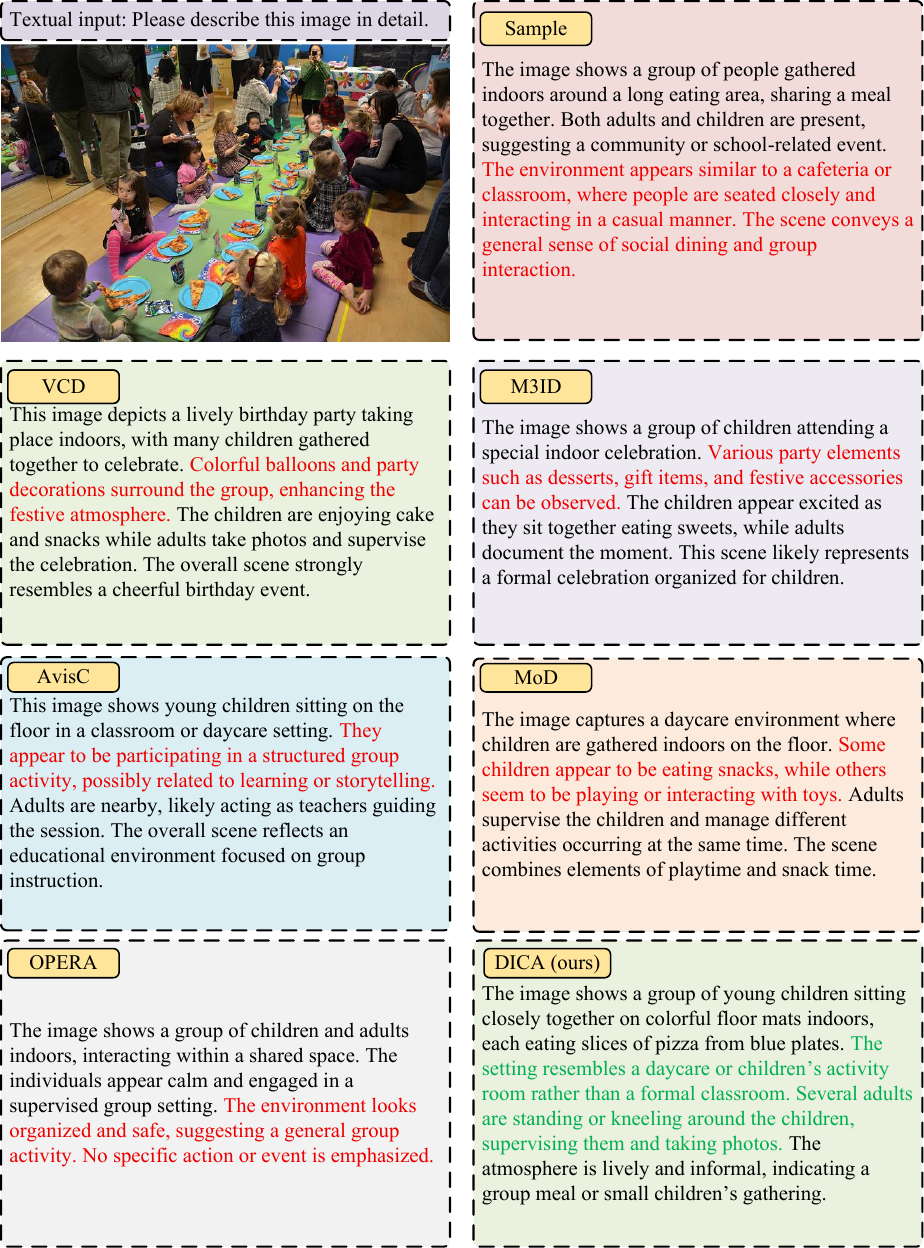}
    \caption{Case study of generative on CHAIR. We compare responses generated by sampling, VCD, M3ID, AvisC, MoD, and OPERA. Hallucinated content is highlighted \textcolor{red}{red}, while more detailed and accurate content is marked in \textcolor{green}{green}.}
    \label{fig:case_1}
\end{figure*}
Figure~\ref{fig:case_1} presents a qualitative comparison of different decoding and alignment methods on a representative indoor scene involving children and adults sharing food. As shown, baseline methods such as VCD, M3ID, VCD, and OPERA tend to introduce inconsistent or hallucinated details when generating descriptions, for example, misidentifying the event as a birthday party, adding non-existent decorations, or over-specifying social activities that are not visually grounded. These errors indicate that the models are overly influenced by spurious correlations or high-level priors rather than the actual visual evidence. In contrast, DICA produces a more faithful and conservative description that remains closely aligned with the image content, avoiding unsupported assumptions while preserving key visual elements. This case study demonstrates that DICA effectively mitigates attention drift and reduces semantic hallucination by correcting intermediate representations, leading to more reliable image-grounded generation.
\begin{figure*}[t]
    \centering

    \includegraphics[width=1.0\linewidth]{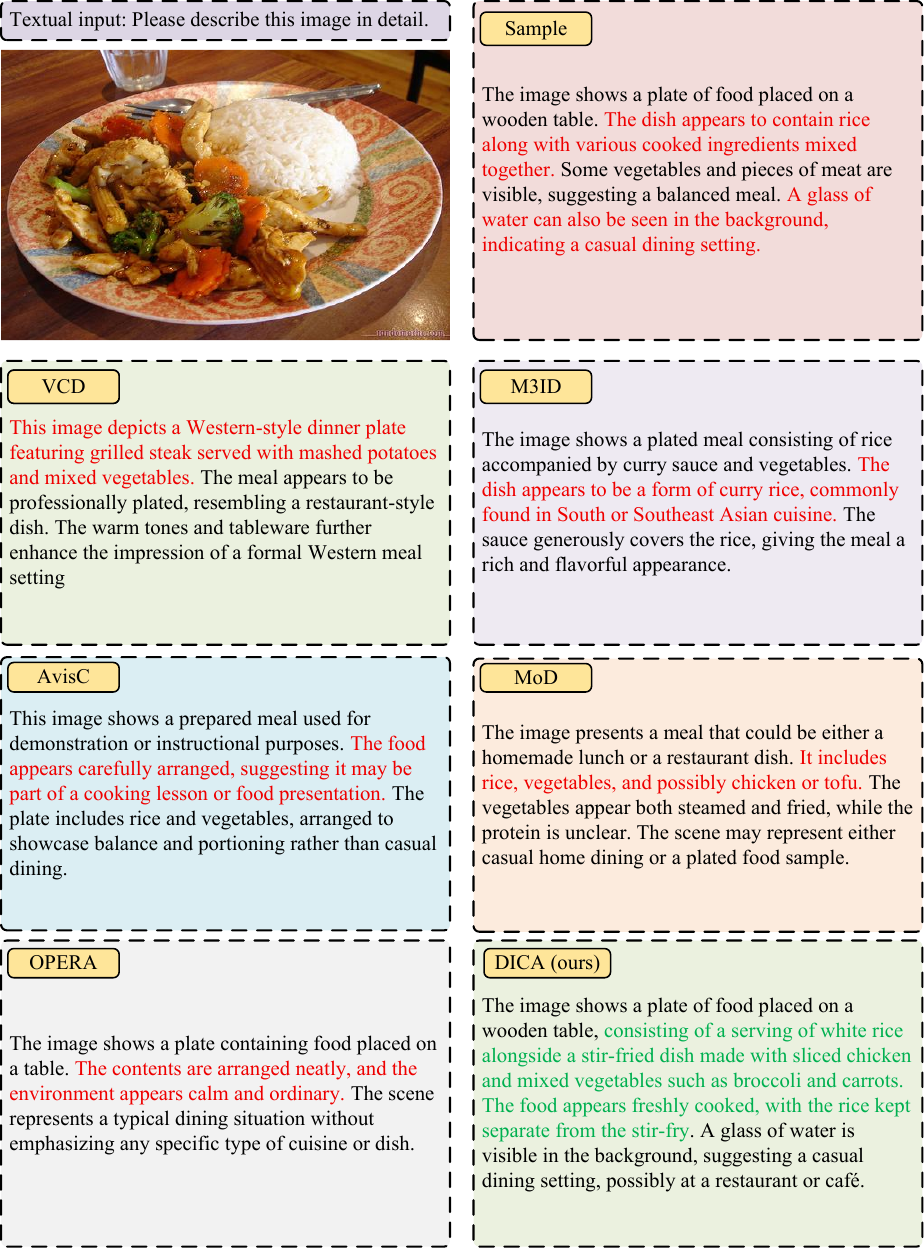}
    \caption{Case study of generative on CHAIR. We compare responses generated by sampling, VCD, M3ID, AvisC, MoD, and OPERA. Hallucinated content is highlighted \textcolor{red}{red}, while more detailed and accurate content is marked in \textcolor{green}{green}.}
        \label{fig:case_2}
\end{figure*}

Figure~\ref{fig:case_2} presents a qualitative comparison of image description results generated by different decoding methods on the same input image. As shown, baseline methods tend to introduce varying degrees of hallucination or over-interpretation. Some outputs incorrectly infer specific cuisines, misidentify ingredients, or add contextual assumptions such as restaurant settings or instructional purposes that are not supported by the visual evidence. These errors indicate that the models are overly influenced by language priors, leading to semantic drift during decoding. In contrast, DICA produces a more faithful and conservative description, accurately reflecting the observable content in the image without introducing unsupported details. This case study highlights the effectiveness of DICA in suppressing hallucinated information and improving visual grounding.
\end{document}